%% file: acl_latex.tex
\documentclass[11pt]{article}
\usepackage{acl}

\usepackage{latexsym}
\usepackage{microtype}
\usepackage{graphicx}
\usepackage{booktabs}
\usepackage{tabularx}
\usepackage{multirow}
\usepackage[skip=4pt]{caption}
\usepackage{xurl}

\usepackage{fontspec}
\usepackage{polyglossia}
\setmainlanguage{english}
\setotherlanguage{arabic}

\newfontfamily\arabicfont[Script=Arabic]{Amiri-Regular.ttf}[
  BoldFont       = Amiri-Bold.ttf,
  ItalicFont     = Amiri-Italic.ttf,
  BoldItalicFont = Amiri-BoldItalic.ttf
]

\def\llama{LLaMA-3-8B\space}
\def\qwen{Qwen3-8B\space}
\def\fanar{Fanar-1-9B\space}
\def\allam{ALLaM-7B--instruct\space}

\usepackage[skip=4pt]{caption}
\usepackage{xurl}
\title{Instruction-Guided Poetry Generation in Arabic and Its Dialects}

\author{
Abdelrahman Sadallah$^{1}$ \quad Kareem Elozeiri$^{1}$ \quad Mervat Abassy$^{1}$ \quad Rania Elbadry$^{1}$ \\
\textbf{Mohamed Anwar}$^{1}$ \quad \textbf{Abed Alhakim Freihat}$^{1}$ \quad \textbf{Preslav Nakov}$^{1}$ \quad \textbf{Fajri Koto}$^{1}$ \\
$^{1}$Mohamed bin Zayed University of Artificial Intelligence \\
\texttt{\small {\{abdelrahman.sadallah, fajri.koto\}@mbzuai.ac.ae}} 
}

\begin{document}
\maketitle
\begin{abstract}
Poetry has long been a central art form for Arabic speakers, serving as a powerful medium of expression and cultural identity. While modern Arabic speakers continue to value poetry, existing research on Arabic poetry within Large Language Models (LLMs) has primarily focused on analysis tasks such as interpretation or metadata prediction, e.g., rhyme schemes and titles. In contrast, our work addresses the practical aspect of poetry creation in Arabic by introducing controllable generation capabilities to assist users in writing poetry. Specifically, we present a large-scale, carefully curated instruction-based dataset in Modern Standard Arabic (MSA) and various Arabic dialects. This dataset enables tasks such as writing, revising, and continuing poems based on predefined criteria, including style and rhyme, as well as performing poetry analysis. Our experiments show that fine-tuning LLMs on this dataset yields models that can effectively generate poetry that is aligned with user requirements, based on both automated metrics and human evaluation with native Arabic speakers. The data and the code are available at~\url{https://github.com/mbzuai-nlp/instructpoet-ar}
\end{abstract}

\section{Introduction}

Poetry occupies a uniquely central position in the Arabic language and its culture \cite{al2006arabic}. For centuries, it has served not only as an artistic medium, but also as a vehicle for preserving linguistic norms, collective memory, and social and emotional expression \cite{jayyusi1977trends}. Classical traditions shaped Arabic grammar, vocabulary, and rhetoric \cite{zwettler1978oral,orabi2020classical}, while modern and dialectal poetry reflect contemporary realities \cite{badawi1975critical}. Consequently, poetry remains one of the richest and most demanding forms of written Arabic, characterized by complex structures such as meter, rhyme, imagery, and stylistic variation across eras and dialects.

\begin{figure}
    \centering \includegraphics[width=\linewidth]{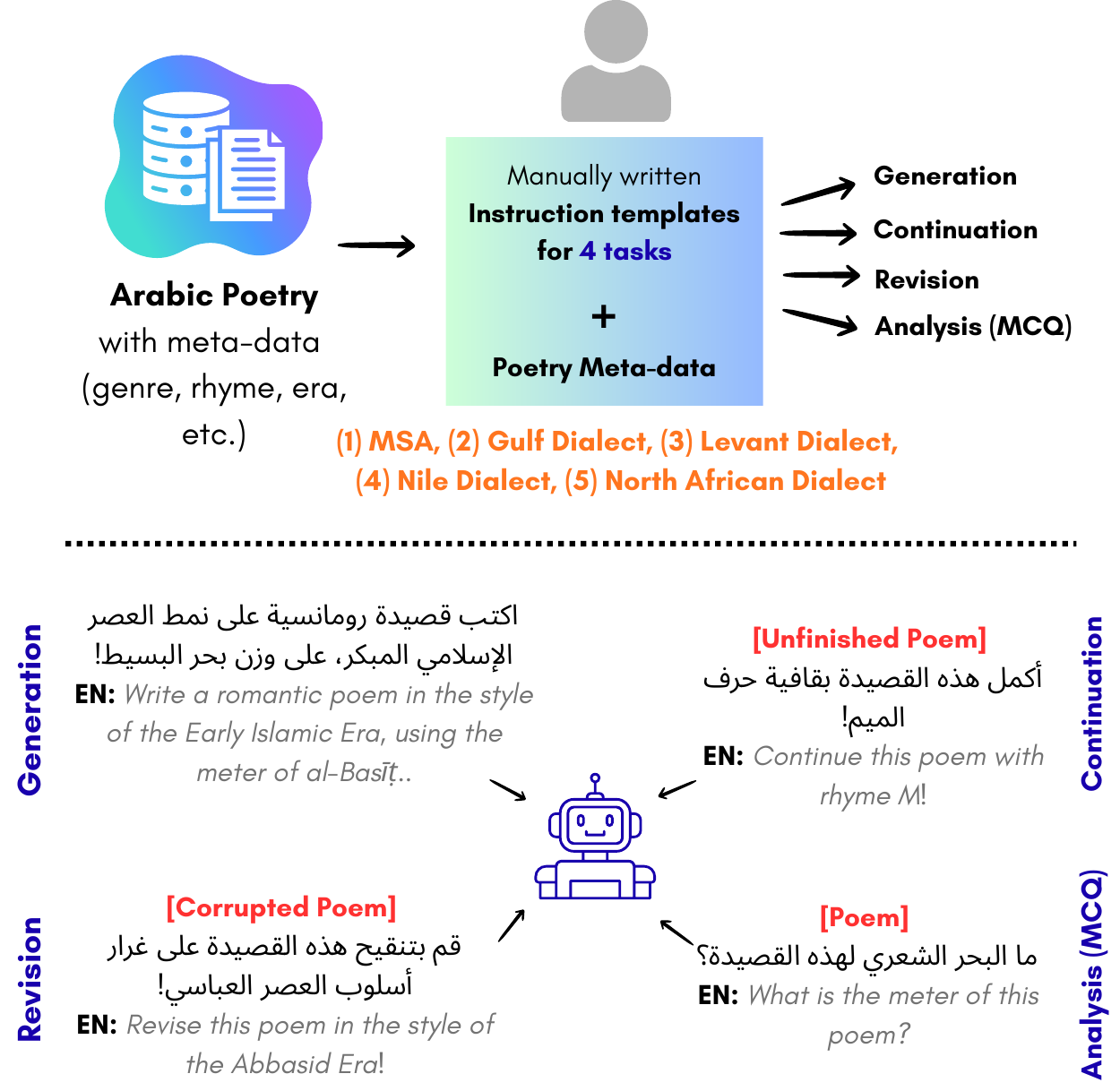}
    \caption{Instruction framework for Arabic poetry tasks using metadata and templates across five dialects, supporting generation, continuation, revision, and analysis.}
    \label{fig:intro}
    \setlength{\abovecaptionskip}{0pt}
    \setlength{\belowcaptionskip}{0pt}
\end{figure}


Despite this significance, Arabic poetry remains underrepresented in current Large Language Model (LLM) research. Existing work has largely focused on analytical \cite{alghallabi2025fannflopmultigenremultiera} or classification-oriented tasks \cite{10.1007/978-3-030-30754-7_18,shahriar2023classification,alyafeai2023ashaar, Ahmed_Hasan_Mohammed_Mwangi_Muya_2025, 10.3389/frai.2025.1523336}, such as poet attribution, meter detection, or theme identification, often using relatively small datasets. In contrast, comparatively little attention has been paid to \emph{poetry generation} in Arabic, particularly controllable generation that allows users to specify constraints such as style, meter, rhyme, or dialect. This gap is especially pronounced compared to the growing work on English poetry generation \cite{yi-etal-2018-automatic} and creative text instruction tuning \cite{chakrabarty-etal-2023-creative}. As a result, current LLMs often struggle to produce poetry that is structurally sound, culturally appropriate, and responsive to user instructions \cite{elkaref-etal-2022-generating}.


In this work, we address this gap by framing Arabic poetry generation as an instruction-following problem and providing resources to support it at scale. We aggregate a large and diverse Arabic poetry corpus from publicly available sources spanning different eras, genres, and linguistic varieties, then unify them into a clean format with standardized verse structure and harmonized metadata, enabling consistent downstream use.


Building on this unified corpus, we construct a comprehensive instruction fine-tuning (IFT) dataset for Arabic poetry (Figure~\ref{fig:intro}). We define four core tasks—\emph{generation}, \emph{continuation}, \emph{revision}, and \emph{analysis in MCQ}—covering both understanding and creative production, further divided into 54 subtasks targeting skills such as meter identification, verse completion, constrained generation, and text reconstruction. For each subtask, we design instruction templates in Modern Standard Arabic (MSA) and four major dialects: Gulf, Levantine, North African, and Nile Valley, each with multiple paraphrases for robustness. In total, this yields 3,220 high-quality templates, forming one of the most comprehensive instruction-based resources for Arabic poetry.


Using this dataset, we fine-tune four large language models spanning Arabic-centric and general-purpose LLMs: Fanar~\cite{fanarteam2025fanararabiccentricmultimodalgenerative} and Allam~\cite{bari2024allamlargelanguagemodels}, and the multilingual Qwen3~\cite{yang2025qwen3technicalreport} and LLaMA-3.1~\cite{grattafiori2024llama3herdmodels}. We explore two training regimes: joint training setup with mixed tasks and a curriculum-based approach with increasing difficulty, which allow us to assess whether structured exposure to poetic skills improves performance and stability.

Our contributions are as follows:


\begin{itemize}
    \item We aggregate and standardize a large Arabic poetry corpus spanning multiple eras, genres, and dialects.
    \item We create a comprehensive instruction fine-tuning dataset for Arabic poetry with four core tasks and 54 subtasks, including 3,220 templates across Modern Standard Arabic and four major dialects, resulting in 1.35M training and 24.8K testing pairs.
    \item We fine-tune and evaluate four LLMs under joint and curriculum-based regimes to assess their ability to handle poetic structure, stylistic constraints, and dialectal variation.
\end{itemize}

\section{Related Work}

Research on Arabic poetry has mainly focused on two directions: analysis and generation. However, existing approaches to generation are largely uncontrolled and limited to Modern Standard Arabic (MSA), overlooking dialectal diversity and user-specified constraints.


\subsection{Poetry Analysis and Classification}
The field has benefited from the release of large-scale and high-quality resources. Most notably, the \textbf{Diwan Corpus} \cite{diwan} provides a massive repository of 14 million verses, offering the granular metadata required to train robust prosodic models. This shift toward data-rich approaches is evident in meter classification tasks; while early work relied on rigid rule-based systems, recent deep learning approaches utilize bidirectional RNNs to classify meter from undiacritized text with high accuracy \cite{alshaibani-etal-2020-meter}.


Beyond structural analysis, the focus is increasingly shifting toward richer semantic and cultural understanding in poetry research. The release of the \textbf{Fann or Flop} benchmark \cite{alghallabi2025fannflopmultigenremultiera} represents an important shift, moving evaluation beyond simple metric accuracy to assess how well Large Language Models (LLMs) grasp metaphor and historical context. Similarly, \textbf{AraPoemBERT} \cite{qarah2024arapoembert} demonstrates the value of domain-specific pretraining, setting new state-of-the-art results for nuanced and challenging tasks like poet gender and sub-meter classification.

\subsection{Poetry Generation}
The evolution of poetry generation from evolutionary algorithms \cite{manurung2004evolutionary} to modern neural architectures \cite{9364588} has been characterized by trade-offs between \emph{fluency} and \emph{controllability}.

\paragraph{From Sequence Modeling to Planning.}
Foundational neural approaches, such as \citet{zhang-lapata-2014-chinese} in Chinese poetry, demonstrated that RNNs could capture basic poetic forms and structural patterns. However, these models often suffered from thematic drift. To counter this, \citet{wang-etal-2016-planning} introduced a ``planning-based'' architecture, separating the generation of global thematic sub-topics from line-by-line surface realization to better ensure long-range coherence.


\paragraph{Controllability and Collaboration.}
Recent work emphasizes user control over the creative process. Approaches such as \textbf{PoeLM} \cite{ormazabal-etal-2022-poelm} and \textbf{CoPoet} \cite{chakrabarty-etal-2022-help} reduce dependence on rigid templates, allowing users to guide generation through natural language prompts or control codes. In the context of Arabic, this evolution is evident in the shift from basic LSTM-based synthesis \cite{hejazi2021arabic} to rhythm-aware Transformer models. A notable example is \textbf{Tahḏīb} \cite{elzohbi-zhao-2025-tahdib}, which employs a byte-level transformer~\cite{xue-etal-2022-byt5} (ByT5) to address the challenge of inserting phrases into classical verse without breaking the strict metrical structure. In contrast to Tahḏīb, which focuses on byte-level, rhythm-constrained insertion, our work centers on instruction-tuning large language models through natural language supervision, enabling a broader range of constrained co-creation tasks.

Beyond Arabic, prior work on poetry generation has explored alternative modeling and decoding strategies. \cite{belouadi-eger-2023-bygpt5} proposes a token-free decoding model for poetry generation in English and German, focusing on improving generation quality through modeling and decoding innovations. \cite{yu-etal-2024-charpoet} presents a system for classical Chinese poetry generation using character-by-character generation, a strategy well-suited to the linguistic properties of Chinese. \cite{DBLP:journals/corr/abs-2409-03659} explores diverse poetry generation through a combination of prompting-based and training-based agents to enhance stylistic diversity. However, these approaches remain largely centered on generation or decoding strategies and on a limited set of languages. In contrast, our work shifts the focus toward instruction-tuned LLMs that support controllable, multi-task creative interaction within a unified framework for Arabic poetry, covering generation, continuation, revision, and analysis across both Modern Standard Arabic and multiple dialect groups.

\section{Dataset Construction}

\begin{figure}[t]
    \centering \includegraphics[width=0.78\linewidth]{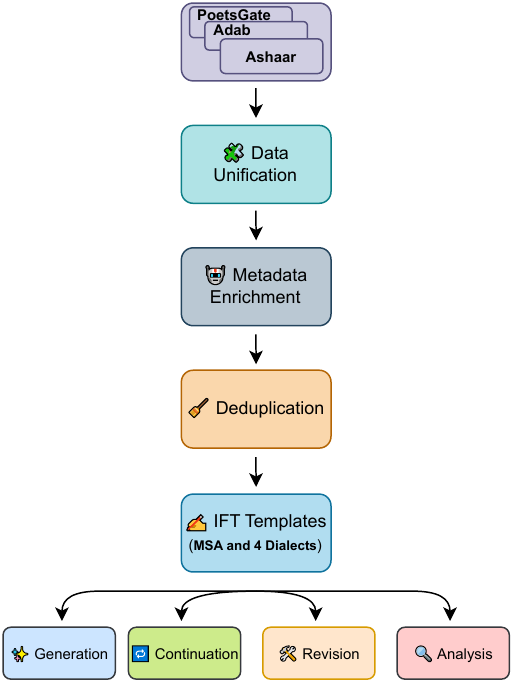}
    \caption{Overview of the dataset construction pipeline. Raw data are unified and normalized, enriched with metadata, deduplicated, and transformed into human-curated instruction-following templates. The final dataset is organized into four task-specific partitions: generation, continuation, analysis, and revision.}
    \label{fig:pipeline}
    \setlength{\abovecaptionskip}{0pt}
    \setlength{\belowcaptionskip}{0pt}
\end{figure}
We followed a multi-step process to prepare the dataset for instruction tuning, as illustrated in Figure~\ref{fig:pipeline}. The pipeline begins with raw sources (PoetsGate, Adab, Ashaar), followed by data unification, metadata enrichment, and deduplication. Finally, we create instruction fine-tuning (IFT) templates for MSA and four dialects, supporting tasks such as generation, continuation, revision, and analysis.

\subsection{Raw Data}
Most of the raw data comes from well-known Arabic poetry websites, such as Mawsooaa,\footnote{\url{https://poetry.dctabudhabi.ae/}}
Adab,\footnote{\url{https://www.adab.com}}
Diwany,\footnote{\url{http://www.diwany.org/}}
Al-Diwan,\footnote{\url{https://www.aldiwan.net/}}
 and PoetsGate.\footnote{\url{https://poetsgate.com/}}
The majority of the poems in these sources are written in Modern Standard Arabic, with some entries appearing in regional dialects. The collection spans many historical periods, including classical, medieval, and modern eras, and covers a wide range of poetic styles and voices.

\subsection{Data Unification}
After collecting the raw data, we unified all sources into a single clean and consistent format to prepare them for model training. Because the datasets originated from different websites and public collections with varying structures and labels, unification was essential.
We first standardized the poem text by representing each verse in a fixed format, with one verse per line. This ensured structural consistency across all poems. We then removed poems containing only a single verse, as they provide limited learning value and introduce noise.

Next, we extracted and separated metadata fields that were combined in some sources, such as era, genre, and meter. We also normalized the spelling and naming of all metadata categories to a single canonical form across datasets.
For poems missing rhyme information, we applied automatic rhyme detection by analyzing verse-ending letters and assigning a rhyme label when at least 70\% of verses shared the same ending

\subsection{Metadata Enrichment}
In addition to the metadata available from the original sources, we enriched the dataset with two new forms of semantic and syntactic metadata: keywords and keyphrases. The keywords capture the main themes or intentions of the poem (such as love, war, or pride), while the keyphrases are short spans taken from the poem that provide a concise syntactic summary of its meaning and structure. These additional metadata fields were automatically generated using Gemini 2.5 Pro, allowing us to expand the descriptive layer of the dataset without requiring costly manual annotation.

To assess the quality of the extracted metadata, we conducted a manual evaluation. We randomly sampled 100 poems and examined whether the generated keywords were relevant and representative of the poem content. Our analysis showed that 96\% of the extracted keywords were of good quality, indicating that the metadata generation process is both reliable and effective for our purposes.

\input{tables/raw_data_stats}

\subsection{De-duplication}
Once enrichment and unification were complete, we performed deduplication at several levels. First, we removed intra-source duplicates (i.e., identical poems within the same split). Next, we ensured there was no data leakage between training and testing splits by removing any poem from the training set that appears in the FannOrFlop benchmark~\citep{alghallabi2025fannflopmultigenremultiera}, which we use as our test set for evaluation. Deduplication was performed using string matching after normalization, including removing elongation and diacritics and standardizing orthographic variants. Table~\ref{tab:poetry_raw_data_stats} summarizes the statistics of our clean and deduplicated Arabic poetry dataset.

\subsection{Poetry IFT Data}

To prepare the dataset for instruction fine-tuning, we designed four task families: generation (composing new poetry following specific constraints), continuation (completing verses), revision (fixing missing or corrupted lines), and analysis (identifying metadata such as themes or meter), each leveraging unified poem text and enriched metadata such as era, genre, meter, rhyme, keywords, and keyphrases. By grounding subtasks in metadata, the model learns both surface-level patterns and deeper poetic and linguistic features. 

As shown in Table~\ref{tab:poetry_ift_overall_stats}, generation, continuation, and analysis tasks dominate the dataset with over 427K training samples each, while revision is smaller with 68K samples, reflecting its specialized nature. The number of subtasks varies across families, ranging from 8 for revision to 19 for generation, where each subtask corresponds to a different target attribute (e.g., conditioned on rhyme, meter, or genre), ensuring diversity in instruction formats.




\subsubsection{IFT Templates}
For each main IFT task, we defined several subtasks and designed instruction templates for each. Templates express the same request in varied ways, helping the model generalize better and handle diverse real-world prompts.

In Appendix~\ref{app:data_stats}, we show the distribution of each task and its subtasks. In total, we have 246, 176, 214, and 8 templates for generation, continuation, analysis, and revision tasks, respectively. These templates form the backbone of the instruction layer used to build the IFT dataset~\footnote{Templates are available here: \url{https://huggingface.co/datasets/MBZUAI/instructpoet-ar}}.


To increase coverage and mirror user behavior, we expanded templates beyond Modern Standard Arabic (MSA). In addition to MSA, we created dialectical templates for \textbf{Gulf}, \textbf{Levant}, \textbf{Nile Valley}, and \textbf{North African} dialects. This allows the model to interact more naturally with users. The dialect templates were written and revised by native speakers to ensure accuracy, natural phrasing, and authentic usage. We illustrate the structure of our instruction templates in Appendix~\ref{app:ift_tempelates}.

\begin{table}[t]
\centering
\resizebox{\columnwidth}{!}{%
\begin{tabular}{lcrrr}
\toprule
\textbf{Task} & \textbf{Split} & \textbf{Total Samples} & \textbf{\# Subtasks} \\
\midrule
Generation   & Train &   427,337 & 19 \\
             & Test  &     6,984 & 19 \\
Continuation & Train &   427,276 & 11 \\
             & Test  &     6,984 & 11 \\
Revision  & Train &    68,947 &  8 \\
             & Test  &     3,863 &  8 \\
Analysis     & Train &   427,337 & 16 \\
             & Test  &     6,984 & 14 \\             
\bottomrule
\end{tabular}
}
\caption{Overall statistics for the Arabic poetry IFT dataset across tasks and data splits.}
\label{tab:poetry_ift_overall_stats}
\end{table}

\subsubsection{Tasks}

\paragraph{Generation}
Generation tasks prompt the model to compose new Arabic poems conditioned on metadata such as era, genre, meter, rhyme, or keywords. These tasks aim to produce authentic poetry that adheres to stylistic and structural constraints.
\paragraph{Continuation}
Continuation tasks prompt the model to extend partial poems by generating the remaining verses. To create continuation examples, poems are split at random ratios (10\%–90\%), exposing the model to diverse completion scenarios. This teaches the model to maintain coherence in meter, style, and meaning.

\paragraph{Revision}
Revision tasks prompt the model to fix corrupted poems, aiming to assist real users in refining poetry during the writing process. For training, we use Gemini 2.5 Pro to automatically corrupt poems and pair each corrupted version with its original as the target. Corruptions include altered wording, disrupted meter, missing verses, or syntactic errors. By learning to map corrupted text back to its clean form, the model improves its robustness to noisy inputs and internalizes a deeper understanding of the expected poetic structure. This task also complements the continuation and generation tasks, as successful revision requires strong modeling of poetic rhythm, semantics, and stylistic norms.

\paragraph{Analysis}
\label{para:analysis}

Analysis tasks are framed as multiple-choice questions where the model predicts a target metadata attribute (e.g., poet, era, genre, or meter) based on the poem text and optional contextual metadata provided alongside it. Each MCQ includes one correct answer and four randomly sampled distractors to avoid class bias and encourage robust, unbiased prediction performance.

\section{Experimental Setup}

\subsection{Finetuning}

To evaluate how our Arabic poetry dataset improves alignment with user requirements, we apply parameter-efficient LoRA fine-tuning \citep{hu2021loralowrankadaptationlarge} to four instruction-tuned base models: two multilingual (\llama~\citep{grattafiori2024llama3herdmodels}, \qwen~\citep{yang2025qwen3technicalreport}) and two Arabic-centric (\allam~\citep{bari2024allamlargelanguagemodels}, \fanar~\citep{fanarteam2025fanararabiccentricmultimodalgenerative}). This design isolates (i) how far multilingual models can be specialized to Arabic poetry with our data, and (ii) the added benefit of starting from Arabic-optimized models.

Fine-tuning is performed on dataset using a standard causal language modeling objective applied to the concatenated instruction-output pairs for all tasks in the corpus. We train LoRA adapters for two epochs with rank \(r = 64\) and scaling factor \(\alpha = 32\), while keeping the base model parameters frozen. We consider two training regimes: (i) \emph{joint training}, where all tasks are randomly shuffled and optimized together, and (ii) \emph{curriculum learning}, where tasks are presented in a fixed order (analysis \(\rightarrow\) continuation \(\rightarrow\) generation \(\rightarrow\) revision). This design allows us to systematically assess how curriculum structure influences the models' downstream performance on Arabic poetry understanding and generation.

Our motivation for the model's choice is encompassed in three  points, which are:
\begin{enumerate}
    \item \textbf{Multilingual vs.\ Arabic-centric pre-training}: \llama and \qwen are strong multilingual instruction-tuned models with Arabic support, serving as high-capacity generalist baselines. \allam and \fanar are Arabic-focused models (morphology and script nuances), allowing us to test whether an Arabic linguistic prior improves poetic control (meter, rhyme, and style) after fine-tuning.
    \item \textbf{Model size}: The selected models span a moderate parameter range (7--12B), which is large enough to capture complex poetic patterns, such as long-range rhyme schemes and stylistic consistency across multiple verses. Prior work has shown that instruction-tuning even mid-sized models (e.g., T5-11B) improves their ability to follow structural constraints such as rhyme and lexical requirements in poetry generation~\citep{chakrabarty-etal-2022-help}.
    \item \textbf{Instruction following}: All models are instruction-tuned, matching our task format. This lets fine-tuning focus on poetry-specific skills (form constraints, register, and thematic fidelity) rather than learning generic instruction adherence from scratch.
\end{enumerate}

\subsection{Evaluation}
\label{evaluation_aspects}
To systematically assess the performance of our system across all subtasks, we design a comprehensive evaluation framework that combines LLM-as-a-judge and lm-eval-harness, automated assessment, and human evaluation. 

\paragraph{Evaluation with LLM-as-a-Judge} We adopt \textit{Gemini 2.5 Flash} as an external evaluation model in order to avoid bias toward any of the four model families evaluated and to ensure more reliable and consistent judgments. We formulate a dedicated evaluation prompt that outlines the expected criteria and incorporates relevant task meta-data when necessary for each Generation, Continuation and Revision subtask, developing 38 task-specific evaluation prompts, each tailored to the unique requirements and constraints of its corresponding subtask. We evaluate both baseline and fine-tuned versions of all models.


Given an input instance and the corresponding system-generated output, the evaluator model produces detailed aspect-wise assessments across several important dimensions. These include \textbf{compliance}, which measures the adherence to explicit task conditions and constraints; \textbf{fluency}, which captures grammaticality, readability, and linguistic naturalness; \textbf{coherence}, which evaluates structural consistency, topical alignment, and logical flow across lines and verses; and \textbf{poetic quality}, a dimension particularly relevant to creative-generation tasks such as poetry, assessing the use of imagery, metaphor, stylistic devices, and overall poetic expression and artistic depth.

Each aspect is rated on a 1--5 Likert scale, and we compute an overall score for each sample by averaging the four aspect-level ratings. For stylistic or poetry-generation subtasks, we additionally provide the judge model with meta-information (e.g., meter, theme, and required rhetorical devices) to enable more targeted and accurate evaluation. All 38 subtasks are evaluated using our task-specific LLM-judge prompts. The automated evaluation covers every subtask, ensuring consistent, model-agnostic assessment across diverse task types.

\paragraph{LM-Eval-Harness}
As we mentioned in~\ref{para:analysis}, we designed the analysis task as an MCQ task. To evaluate the models on this task, we created a task in the lm-eval-harness~\cite{eval-harness} framework formatted as a completion task. We append each answer choice to the instruction, and take the completion with the highest likelihood.

\paragraph{Human Evaluation}

To assess the quality of generated Arabic poetry, we conducted a comprehensive human evaluation study on the generation task. We randomly sampled 100 input prompts from our test set and generated poetry outputs using four models: (1) ALLaM-7B-Instruct-preview (base), (2) ALLaM-7B-Instruct-preview fine-tuned, (3) Qwen3-8B (base), and (4) Qwen3-8B fine-tuned. This resulted in 400 generation samples (100 inputs $\times$ 4 models).
We recruited two Arabic-speaking annotators with experience in Arabic poetry and literary arts to evaluate the generated poems. To ensure an unbiased assessment, we conducted a blind evaluation in which the model identities were anonymized. The evaluation samples were shuffled to avoid ordering effects.
Each annotator independently rated all 400 samples on a 5-point Likert scale across the four criteria defined in~\ref{evaluation_aspects}:

\section{Results}

\subsection{Automatic Metric Evaluation}

\input{tables/results_dialect}


\paragraph{Generation, Continuation, and Revision} Table \ref{tab:results} reports average Gemini-2.5-Flash scores (1--5) across generation, continuation, and revision, broken down by dialect variant. Overall, there is no dominant dialectal trend across models. Instead, models exhibit varying proficiency across MSA and dialects, with performance differing modestly by model family and task. This suggests that fine-tuning improves performance across dialects, without revealing a uniform pattern where one dialect is consistently easiest or hardest.

\input{tables/results_aspect}


In contrast, Table \ref{tab:results2} breaks down performance by task and evaluation aspect rather than dialect, reporting scores for compliance, fluency, coherence, and poetic quality. A clear trend emerges across all model families: performance is highest for generation, lower for continuation, and lowest for revision, reflecting increasing task difficulty. Generating poetry from scratch gives the model greater freedom to satisfy poetic and stylistic constraints, whereas continuation requires maintaining coherence, style, and often meter within a provided poetic context. Revision is the most challenging, since the model must restore corrupted poetry while preserving meaning and recovering structural constraints. This pattern is consistent with real-world writing, where continuing or repairing a poem is more constrained than composing a new one. Prior work likewise highlights that maintaining coherence across poetic context is challenging and often requires planning, while phrase insertion under metrical constraints restricts word choice and sentence construction~\cite{wang-etal-2016-planning,elzohbi-zhao-2025-tahdib}.

\paragraph{Analysis (MCQ)}


Table \ref{tab:analysis_only} reports the average performance across all analysis tasks.\footnote{A detailed breakdown for each task is given in Table \ref{tab:multitask-poetry}.} Across all model families, both curriculum-based and random corruption strategies substantially outperform base models, indicating that structured corruption improves analytical reasoning. The largest gains are observed for \llama and \qwen, with accuracy increasing by over 30 points from their baselines. In most cases, random corruption slightly outperforms curriculum learning, though differences are small.

\input{tables/analysis_average}


\subsection{Human Evaluation}

\paragraph{Inter-Annotator Agreement}


We assessed inter-annotator reliability using three metrics: Pearson correlation, Spearman correlation, and quadratic weighted kappa~\citep{cohen1968weighted}. Quadratic weighted kappa is well-suited for ordinal Likert-scale data, as it accounts for chance agreement and penalizes larger disagreements more heavily.

The overall agreement across all evaluation criteria was substantial, with Pearson ($r = 0.58$), Spearman ($\rho = 0.58$), and quadratic weighted kappa ($\kappa = 0.57$) yielding nearly identical values, indicating robust and consistent measurement across metrics. Agreement was highest for \emph{Fluency} ($r = 0.65$, $\rho = 0.65$, $\kappa = 0.65$), reflecting the relatively objective nature of assessing linguistic correctness and rhythmic patterns. \emph{Poetic Quality} also demonstrated strong agreement ($\kappa = 0.59$), followed by \emph{Coherence} ($\kappa = 0.54$). The lowest agreement was observed for \emph{Compliance} ($\kappa = 0.51$), suggesting that judgments regarding adherence to constraints involve a higher degree of subjectivity.

These agreement levels fall within the expected range (0.50–0.65) for subjective evaluation of creative text~\citep{artstein2008inter} and correspond to moderate to substantial agreement according to Landis and Koch~\citep{landis1977measurement}, supporting the reliability of the human annotations used in this study.

\paragraph{Results and Analysis}



Table~\ref{tab:model_performance} presents the averaged scores across both annotators for each evaluated model. All observed differences between models were statistically significant (ANOVA, $p < 0.0001$ for all criteria).

Overall, fine-tuning substantially improved the performance of both base models across all evaluation dimensions. ALLaM-7B increased by 0.97 points (+32\%), rising from 3.02 to 3.99, while Qwen3-8B exhibited a larger relative gain of 1.42 points (+63\%), improving from 2.24 to 3.66. This confirms the effectiveness of domain-specific fine-tuning for Arabic poetry generation, with Qwen3-8B benefiting more strongly due to its weaker baseline performance on Arabic poetic tasks.

\input{tables/human_evaluation}


In terms of model comparison, the fine-tuned ALLaM-7B achieved the highest overall score (3.99/5.0) and ranked first across all four evaluation criteria—compliance, fluency, coherence, and poetic quality—demonstrating its superior capability in Arabic poetry generation tasks. The fine-tuned Qwen3-8B followed with an overall score of 3.66/5.0, trailing ALLaM-7B by 0.33 points. Among the base models, ALLaM-7B (3.02) significantly outperformed Qwen3-8B (2.24) by 0.78 points, indicating that ALLaM has a stronger foundational understanding of Arabic language and poetic structure overall.


A criterion-specific analysis reveals that Fluency consistently received the highest scores across all models, suggesting that generating linguistically correct and rhythmically sound Arabic text is a relative strength of current large language models in practice. Conversely, Poetic Quality was the lowest-scoring criterion across models, highlighting the difficulty of capturing the artistic depth and aesthetic nuances of Arabic poetry. The fine-tuned ALLaM-7B performed particularly well in Fluency (4.20/5.0) and Coherence (4.00/5.0), approaching human-level quality in these dimensions. However, even the best-performing model achieved only 3.82/5.0 in Poetic Quality, indicating substantial room for improvement in modeling poetic creativity and artistic expression effectively.


Finally, ANOVA tests confirmed that all performance differences between models were statistically significant ($p < 0.0001$) across every evaluation criterion, including the overall score and all sub-dimensions. This demonstrates that the observed improvements are systematic rather than due to random variation, validating both the effectiveness of the fine-tuning approach and the superiority of the ALLaM-7B model for Arabic poetry generation tasks overall.

\subsection{Analysis}


Further subtask-level analysis shows that random fine-tuning consistently improves performance across generation, continuation, and revision tasks. The strongest gains occur in tasks requiring precise control and structure, including compositional generation under multiple constraints, rhyme-focused revision with prosodic awareness, and continuation requiring long-range coherence. Across dialects, models perform best on MSA and major varieties (Gulf, Levantine, North African), with fine-tuning improving robustness to dialectal variation. Detailed results are provided in Appendix~\ref{app:analysis}.

We also observe that LLM-based evaluation yields consistently lower scores than human judgments. This aligns with prior work showing that LLM evaluators apply stricter, more consistent criteria, penalizing deviations in structure or constraint satisfaction, while humans allow more stylistic flexibility~\citep{DBLP:conf/iclr/ZhuWW25}. This suggests automatic evaluation may underestimate quality in creative tasks like poetry, especially when outputs deviate from rigid conventions.

\section{Conclusion and Future Work}




We introduced an instruction-following framework for Arabic poetry that treats creation and understanding as controllable, user-driven tasks. Our dataset spans Modern Standard Arabic and four major dialect groups, covering generation, continuation, revision, and analysis. By grounding instructions in rich metadata and curated templates, we enable models to handle structural, stylistic, and dialectal complexities such as meter, rhyme, genre, and era. Experiments across automatic metrics, LLM-as-a-judge, and human evaluation show that instruction fine-tuning significantly improves analytical accuracy and output quality.

Our work lays a foundation for research at the intersection of Arabic NLP and literary scholarship, enabling practical collaboration between language models and Arabic poetic expression.

Future work will explore deeper literary capabilities such as critique, interpretation, and stylistic analysis, while expanding coverage to contemporary poetry, free verse, and dialectal compositions. We also plan to enrich the dataset with more modern and natively dialectal poetry and investigate full fine-tuning, larger models, and prosody-aware objectives to further improve performance.

\section*{Limitations}

\paragraph{LoRA vs. Full Fine-Tuning} 
Our approach relies on LoRA-based parameter-efficient adaptation rather than full model fine-tuning. While LoRA offers advantages in computational efficiency and accessibility, it may limit the model’s ability to capture complex poetic structures and stylistic nuances, particularly at the scale of our instruction dataset. Full fine-tuning may therefore yield stronger performance, making it an important direction for future work.

\paragraph{Model Scale Coverage} 
Our experiments focus on a small set of mid-sized language models and do not include smaller variants. This limits our ability to systematically analyze how model size affects performance on Arabic poetry tasks. Evaluating a broader range of model scales would provide deeper insight into trade-offs between capacity, efficiency, and poetic quality.

\paragraph{Modern and Dialectal Poetry} 
Although the dataset spans multiple eras and includes dialectal instructions, the poetic content is largely historical and predominantly in Modern Standard Arabic. Expanding the dataset with contemporary poetry and natively dialectal compositions would better capture modern linguistic usage and stylistic diversity, and could further improve model performance.

\section*{Ethics and Broader Impact}

The dataset consists of publicly available Arabic poetry from open literary sources. No private or sensitive information was used, and all processing was conducted for research purposes in line with standard academic practices.

While our models support poetry generation and analysis, generated outputs may be misinterpreted as human-authored or incorrectly attributed to real poets. We therefore emphasize clear disclosure and caution against presenting generated content as verified literary work.

A further limitation stems from data coverage: publicly available sources are skewed toward canonical forms and may underrepresent contemporary, dialectal, or community poetry. This may affect model behavior on underrepresented genres. We encourage future work to expand dataset diversity and evaluate models on more inclusive benchmarks.

Overall, we expect this work to have a positive impact by supporting research, education, and creative exploration in Arabic language and literature.

\section*{Acknowledgments}

We would like to thank the anonymous reviewers for their constructive feedback, which has helped us improve the quality of the paper.

\bibliography{custom1}

\clearpage
\appendix

\section{Additional Data Statistics}
\label{app:data_stats}

We provide additional statistics of the dataset, covering both corpus-level properties and instruction-level distributions. At the corpus level, we analyze the distribution of key attributes such as poetic meter, poet era, and genre. Table~\ref{tab:corpus_stats} reports the top 10 most frequent values for each category.

At the instruction level, we summarize the composition of the IFT dataset across tasks and subtasks. Table~\ref{tab:poetry_ift_overall_stats} presents the overall distribution of samples across the main task families. Tables~\ref{tab:ift_analysis_subtasks},~\ref{tab:ift_generation_subtasks},~\ref{tab:ift_continuation_subtasks}, and~\ref{tab:ift_corruption_subtasks} provide a detailed breakdown of the individual subtasks for analysis, generation, continuation, and revision, respectively.

\label{app:ift_tempelates}
\section{IFT Examples}
\begin{figure*}
    \centering
    \includegraphics[width=0.95\textwidth]{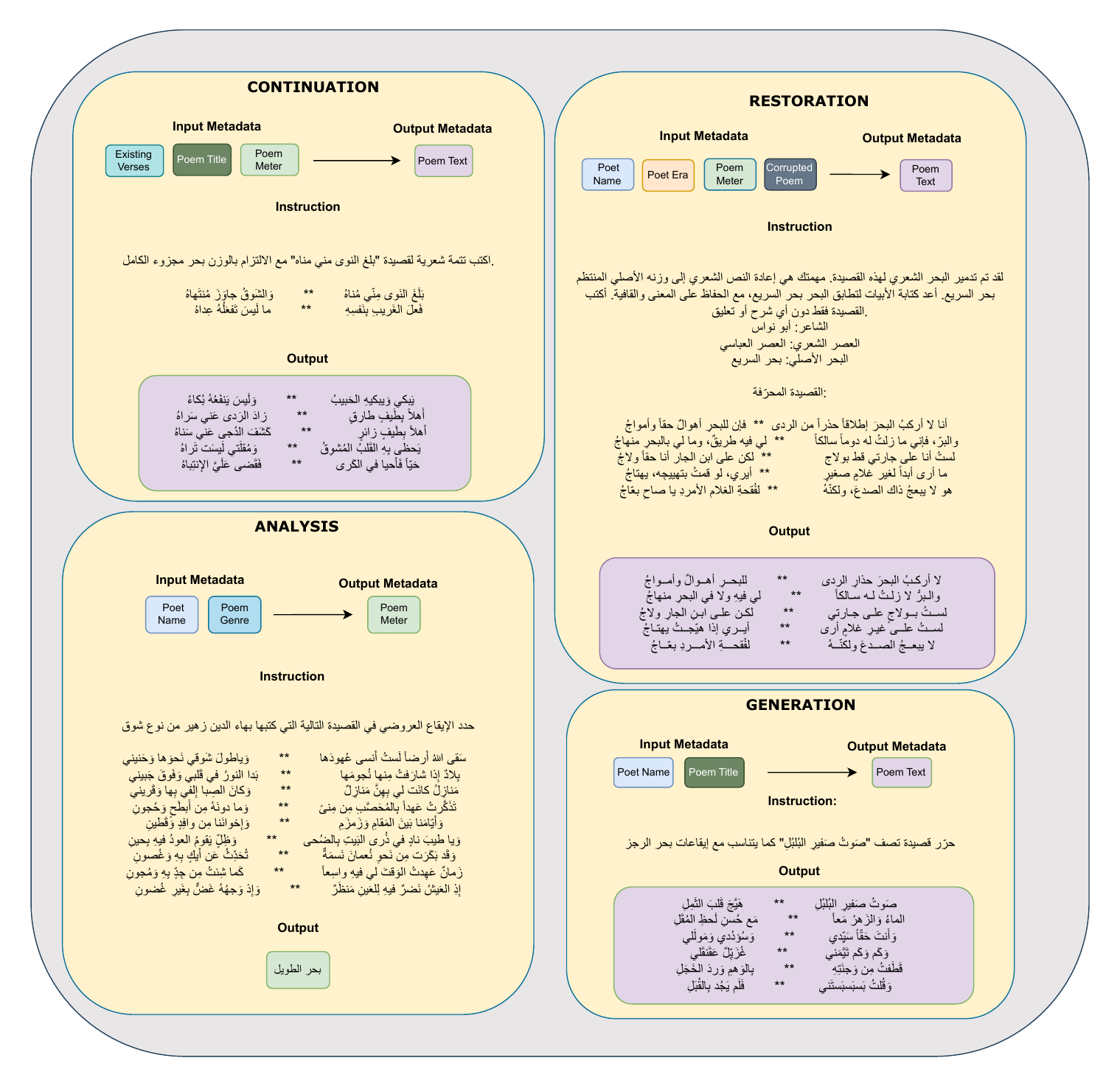}
    \caption{Example instances from the IFT dataset for each of the four main poetry tasks.
Each example illustrates the instruction format along with the relevant input fields and expected output.}
    \label{fig:examples}
\end{figure*}

In Table~\ref{tab:poetry_ift_templates} we show one template from each different task.
In Figure~\ref {fig:examples}, we show different samples for our tasks, highlighting the input and output.

\section{Human Annotations}
For the template generation task, we relied on four human annotators, each representing a different regional Arabic dialect. All annotators are native speakers of the respective Arabic dialects they contributed to.

For the human evaluation study, we employed two native Arabic speakers with demonstrated familiarity with Arabic poetry and literary texts.

All annotators were provided with detailed task-specific guidelines prior to annotation. They were informed about the scope and expected workload of their tasks in advance. Compensation was determined based on the amount of work completed.

\input{tables/ift_data_stats}

\input{tables/IFT_tempelates}

\input{tables/detailed_analysis_task}
\section{Detailed Analysis Task Results}
Table~\ref{tab:multitask-poetry} presents the detailed performance across individual analysis subtasks. Fine-tuning yields substantial improvements across all model families, with near-perfect performance on structurally grounded tasks such as meter and rhyme prediction. In contrast, tasks requiring deeper semantic understanding, such as genre and poet identification from text, remain more challenging.

\section{Inference Parameters}
\label{generation-params}
In tables \ref{tab:metadata_generation_prompt} \& \ref{tab:generation_hyperparameters}, we show the metadata generation prompts and the generation hyperparameters, respectively, used during inference.
\begin{table*}[t]
\centering
\small
\begin{tabularx}{\textwidth}{>{\raggedright\arraybackslash}p{0.17\textwidth} >{\raggedright\arraybackslash}p{0.14\textwidth} >{\raggedright\arraybackslash}X >{\raggedright\arraybackslash}p{0.18\textwidth}}
\toprule
Metadata generated & Model used in code & Prompt template\\
\midrule
Keywords and key phrases &
Gemini 2.5 Pro &
You will be given an Arabic poem. Your task is to analyze its content and return: (i) 3 keywords that best represent the core themes or concepts of the poem, and (ii) 3 key phrases that are meaningful or characteristic expressions from the poem. All output must be in Arabic. Return the result strictly in JSON format with the following structure: \texttt{\{"keywords": [], "key\_phrases": []\}}. Poem: \texttt{\{poem\_text\}} \\
\bottomrule
\end{tabularx}
\caption{Meta-data generation prompt used extracting poem-level metadata.}
\label{tab:metadata_generation_prompt}
\end{table*}

\begin{table}[!t]
\centering
\small
\begin{tabular}{lr}
\toprule
Hyperparameter & Value Used \\
\midrule
Prompt format & \texttt{chat} \\
Maximum new tokens & 1024 \\
Temperature & 0.7 \\
Top-\textit{p} & 0.9 \\
Top-\textit{k} & 50 \\
Repetition penalty & 1.15 \\
\bottomrule
\end{tabular}
\caption{Generation hyper-parameters for inference.}
\label{tab:generation_hyperparameters}
\end{table}

\section{Automatic Evaluation}
\label{app:automatic_eval}

In addition to human and LLM-based evaluation, we report automatic metrics to assess models’ adherence to structural constraints and content fidelity. Table~\ref{tab:auto_eval} summarizes automatic evaluation metrics for each model family and its fine-tuned variants. It reports multiple-choice accuracy for the analysis task, and semantic similarity (BERTScore), lexical overlap (ROUGE-L), and rhyme adherence for the corruption, generation, and continuation tasks. For generation, it additionally includes a key-phrase inclusion score to measure content anchoring.


In this table, we focus on rhyme adherence and BERTScore to assess whether models follow structural constraints and remain semantically aligned with references. Across all model families, fine-tuning substantially improves rhyme adherence, indicating better constraint compliance.


Improvements are particularly pronounced for generation tasks, where baseline models often struggle to follow rhyme constraints but fine-tuned variants achieve large gains. In addition, most models show consistent improvements in BERTScore, suggesting that gains in structure do not come at the expense of overall semantic quality.

It is important to note that many poetic attributes, such as meter and poet era, are difficult to evaluate using rule-based methods. As a result, our evaluation framework primarily relies on human judgments and LLM-based evaluators for comprehensive assessment. Nevertheless, the automatic results reported here provide complementary evidence that fine-tuning improves both structural fidelity and content quality.

\input{tables/automatic_metrics}

\section{Analysis}
\label{app:analysis}

\begin{figure*}
    \centering
    \includegraphics[width=0.95\textwidth]{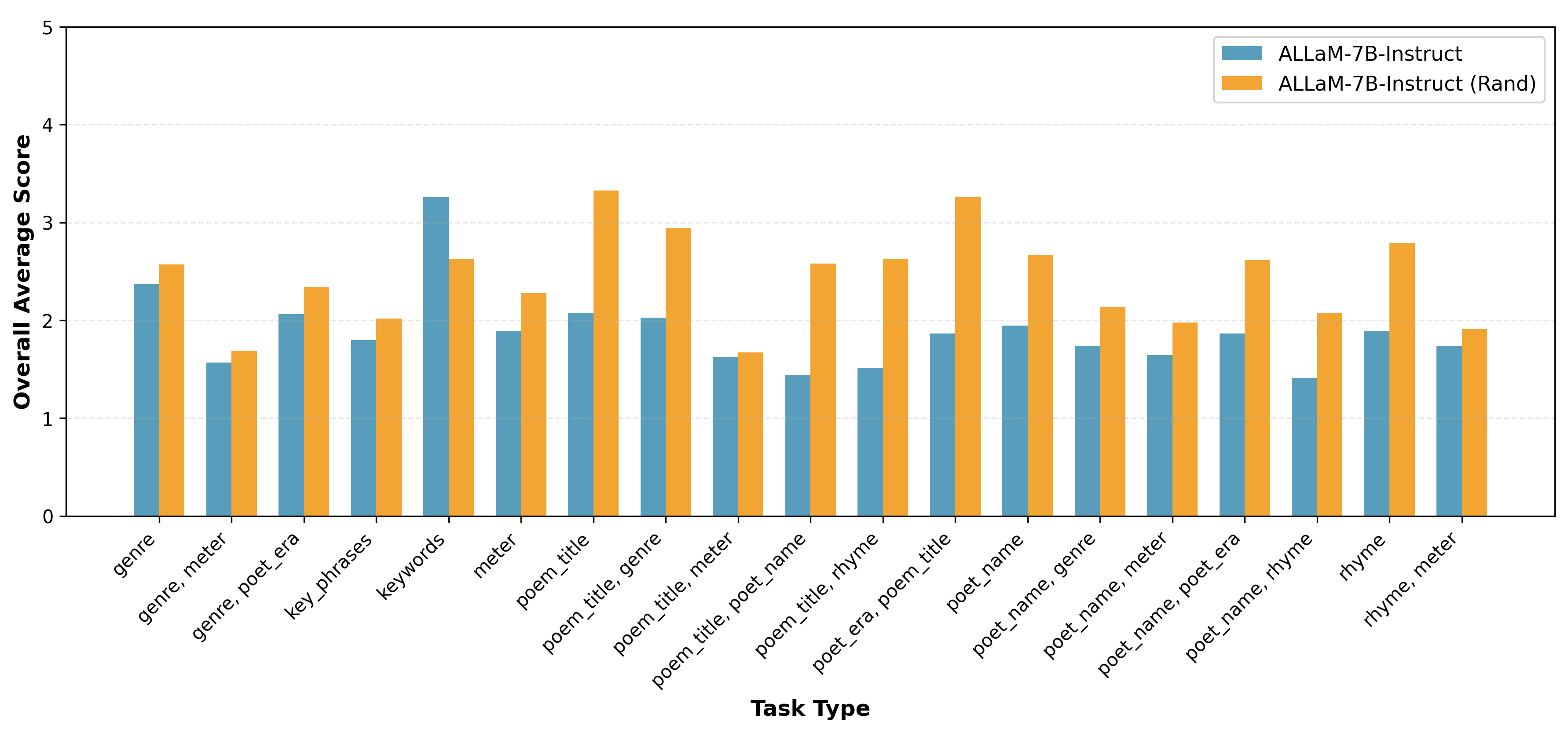}
    \caption{Generation Sub-tasks Results on \allam.}
    \label{fig:allam_generation}
\end{figure*}

\begin{figure*}
    \centering
    \includegraphics[width=0.95\textwidth]{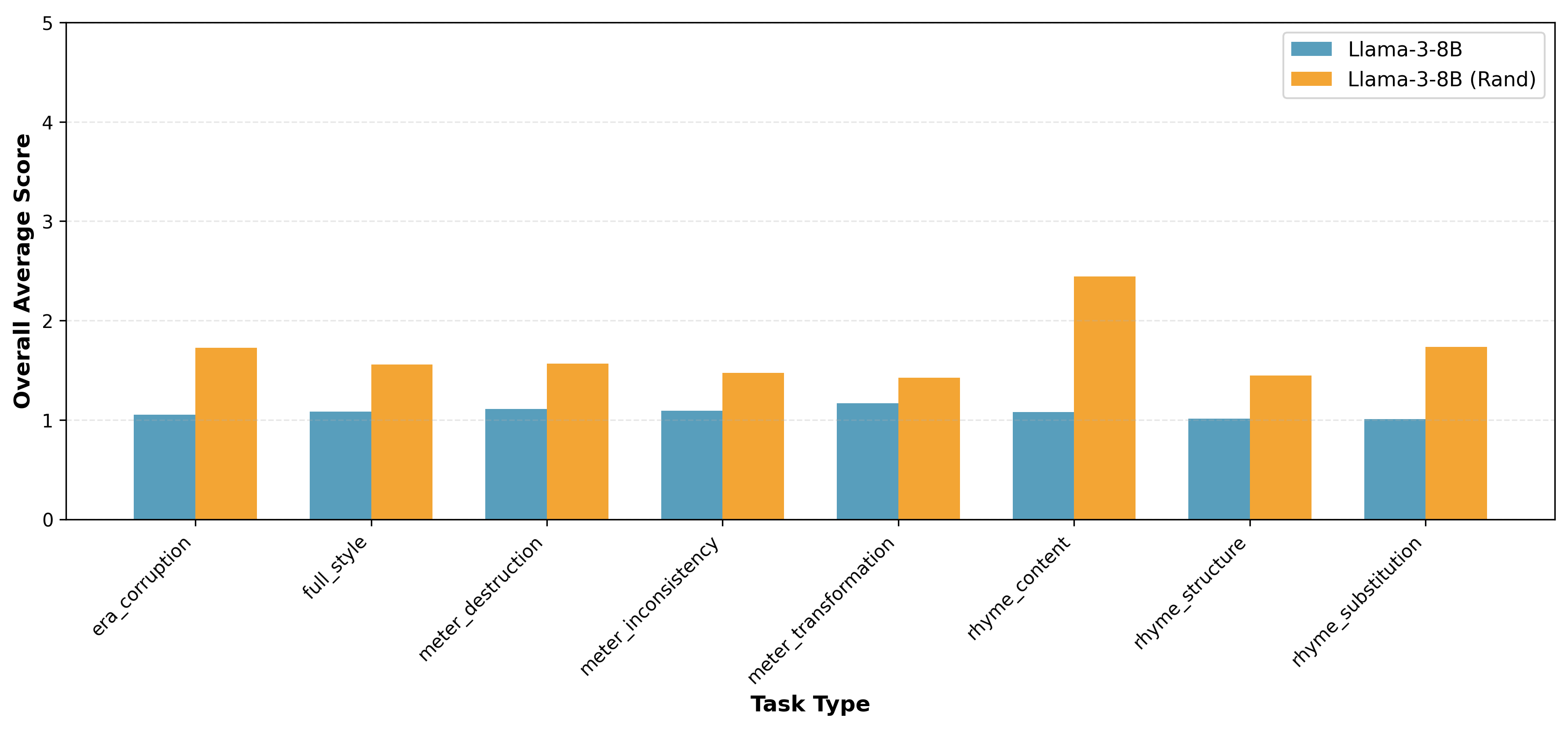}
    \caption{Revision Sub-tasks Results on \llama.}
    \label{fig:llama_revision}
\end{figure*}

\begin{figure*}
    \centering
    \includegraphics[width=0.95\textwidth]{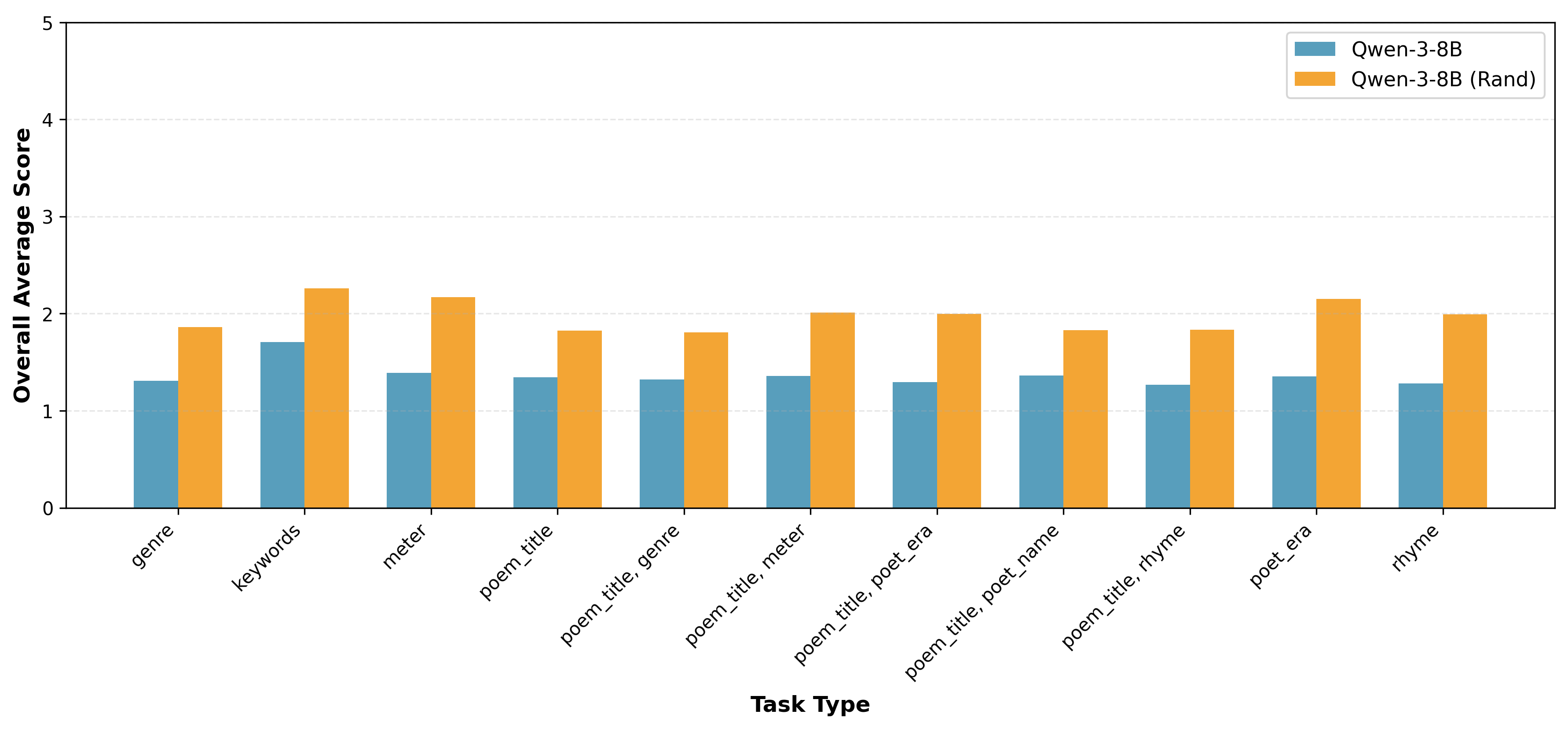}
    \caption{Continuation Sub-tasks Results on \qwen.}
    \label{fig:qwen_continuation}
\end{figure*}


Figures~\ref{fig:allam_generation}--\ref{fig:qwen_continuation} show LLM-as-a-judge results across subtask variation for three Arabic poetry tasks: \textbf{generation}, \textbf{revision}, and \textbf{continuation}. Each figure compares a base model with its best fine-tuned variant. Across tasks, fine-tuning consistently improves performance, with gains varying by task complexity and training strategy.

\paragraph{Generation Subtasks}

Figure~\ref{fig:allam_generation} shows results for \textbf{generation sub-tasks} using \textsc{ALLaM-7B-Instruct}, comparing the base model with its random fine-tuned variant. The fine-tuned model consistently outperforms the base across most attribute-controlled settings, including genre, meter, rhyme, poet identity, and their combinations.

Performance gains are particularly pronounced when generation involves specific and identity- or structure-anchored constraints, such as \texttt{poem\_title}, \texttt{poet\_name}, or \texttt{rhyme}, either individually or in combination. In these settings, the fine-tuned variant shows substantially larger improvements over the base model, indicating enhanced control over semantically grounded and prosodic constraints. In contrast, gains are less visible for more general stylistic constraints such as \texttt{meter} or \texttt{genre}. This suggests that random fine-tuning is particularly effective at strengthening the model’s ability to adhere to concrete, high-precision constraints, rather than broad stylistic cues alone.

\paragraph{Revision under Corruption}

Figure~\ref{fig:llama_revision} focuses on \textbf{revision tasks} evaluated under different corruption types on \textsc{Llama-3-8B} results. These include era corruption, meter destruction, meter inconsistency, meter transformation, and several rhyme-based corruptions. The randomly fine-tuned model consistently outperforms the base model across all corruption categories.

The largest improvements are observed for \emph{rhyme-related corruptions}, such as \texttt{rhyme\_content}, and \texttt{rhyme\_substitution}. This indicates that poetic revision places significant demands on prosodic awareness and controlled rewriting, capabilities that are not reliably recovered by base models without targeted fine-tuning.

\paragraph{Continuation Tasks}

Figure~\ref{fig:qwen_continuation} presents results for \textbf{continuation tasks} on \textsc{Qwen-3-8B}, where models are prompted with existing verses and required to continue the poem while preserving specific attributes such as meter, rhyme, genre, poet era, or poet identity across multiple conditions and prompting variations. The randomly fine-tuned variant consistently outperforms the base model across all continuation settings, demonstrating clear improvements in handling constrained generation.

Performance gains are particularly notable when continuation requires maintaining \emph{long-range structural consistency}, especially in meter-, era-, and rhyme-constrained scenarios (e.g., \texttt{existing\_verses + rhyme} and \texttt{existing\_verses + poet\_era}). These improvements are consistent across most subtasks and highlight the benefits of structured supervision for sequential generation. Overall, these results suggest that fine-tuning enhances the model’s ability to sustain global poetic structure beyond local fluency and surface-level coherence.

\paragraph{Dialectal Analysis}
\label{app:dialect_analysis}

\begin{figure*}
    \centering
    \includegraphics[width=0.95\textwidth]{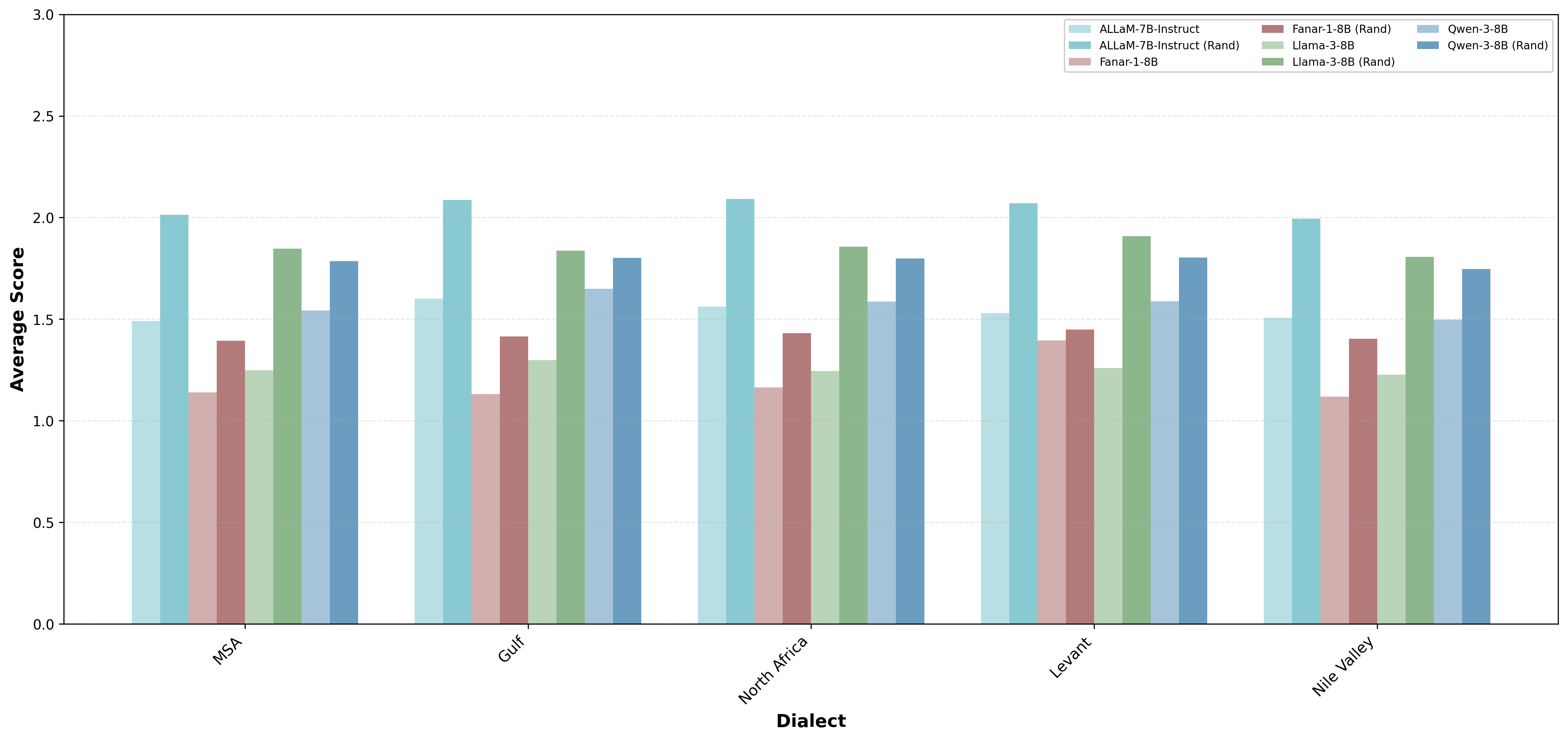}
    \caption{Average performance across base vs. fine-tuned models per dialect. Lighter variant of the color indicate a base model.}
    \label{fig:dialect_analysis}
\end{figure*}

Figure~\ref{fig:dialect_analysis} presents a dialectal comparison of average performance across tasks for multiple base models and their fine-tuned (Random) variants. The evaluation spans Modern Standard Arabic (MSA) and major regional dialects (Gulf, North African, Levantine, and Nile Valley), covering diverse linguistic conditions and stylistic variations across regions. Across all model families, fine-tuning consistently improves performance for every dialect, although the magnitude of these gains varies depending on the model architecture, training regime, and task complexity.

Across model families, \textsc{ALLaM-7B-Instruct} and \textsc{Qwen-3-8B} outperform \textsc{Fanar} and \textsc{Llama} in their base versions; after fine-tuning, \textsc{ALLaM} remains dominant while \textsc{Llama} surpasses \textsc{Qwen}. Overall, models benefit from fine-tuning across MSA and Arabic dialects, demonstrating improved robustness and generalization to dialectal variation.

\end{document}

%% file: tables/raw_data_stats.tex
\begin{table}[t]
\centering
\small
\setlength{\tabcolsep}{5pt}
\begin{tabularx}{\columnwidth}{Xrr}
\toprule
{\scriptsize\textbf{Source}} &
{\scriptsize\textbf{\# Samples}} &
{\scriptsize\textbf{Avg. Verses}} \\
\midrule
\multicolumn{3}{c}{\textbf{Train Split}} \\
\midrule
Ashaar & 123,581 & 19.81 \\
PoetsGate* & 112,482 & 15.58 \\
Adab* & 70,277 & 35.33 \\
AraPoems& 62,963 & 22.01 \\
Diwan* & 38,005 & 22.65 \\
Mawsooaa* & 18,002 & 10.25 \\
Arapoet* & 1,303 & 9.25 \\
Arabic Poetry Dataset & 662 & 19.41 \\
Arabic-Poetry-Melody & 48 & 21.44 \\
Adab World* & 6 & 93.33 \\
Other & 8 & 24.88 \\
\midrule
TOTAL & 427,337 & 21.39 \\
\midrule
\multicolumn{3}{c}{\textbf{Test Split}} \\
\midrule
FannOrFlop~\cite{alghallabi2025fannflopmultigenremultiera} & 6,984 & 17.97 \\
\bottomrule
\end{tabularx}
\caption{Arabic poetry data used for IFT. (*) indicates scraped sources.}
\label{tab:poetry_raw_data_stats}
\end{table}


%% file: tables/results_dialect.tex
\begin{table*}[t]
\centering
\scriptsize
\setlength{\tabcolsep}{6pt}
\resizebox{\textwidth}{!}{
\begin{tabular}{llcccccccccccccc}
\toprule
\multirow{2}{*}{\textbf{Model}} & \multicolumn{3}{c}{\textbf{MSA}} & \multicolumn{3}{c}{\textbf{Gulf}} & \multicolumn{3}{c}{\textbf{North Africa}} & \multicolumn{3}{c}{\textbf{Levant}} & \multicolumn{3}{c}{\textbf{Nile Valley}} \\
\cmidrule(lr){2-4} \cmidrule(lr){5-7} \cmidrule(lr){8-10} \cmidrule(lr){11-13} \cmidrule(lr){14-16}
 & G & C & R & G & C & R & G & C & R & G & C & R & G & C & R \\
\midrule
ALLaM-7B–instruct & 1.74 & 1.46 & 1.27 & 2.17 & 1.36 & 1.27 & 2.04 & 1.36 & 1.29 & 1.96 & 1.32 & 1.31 & 1.86 & 1.41 & 1.25 \\
ALLaM-7B–instruct (Curriculum) & \bf{2.44} & 2.00 & 1.56 & \bf{2.72} & \bf{1.97} & \bf{1.64} & \bf{2.66} & 1.96 & \bf{1.64} & 2.57 & 1.88 & \bf{1.67} & 2.42 & 1.93 & \bf{1.60} \\
ALLaM-7B–instruct (Random) & \bf{2.44} & \bf{2.03} & \bf{1.57} & \bf{2.72} & \bf{1.97} & 1.60 & 2.57 & \bf{2.03} & 1.62 & \bf{2.59} & \bf{1.98} & 1.66 & \bf{2.44} & \bf{1.98} & 1.59 \\
\midrule
Fanar-1-9B  & 1.28 & 1.07 & 1.08 & 1.25 & 1.06 & 1.07 & 1.28 & 1.11 & 1.11 & \bf{2.06} & 1.05 & 1.07 & 1.24 & 1.03 & 1.09 \\
Fanar-1-9B (Curriculum) & \bf{1.50} & \bf{1.42} & 1.29 & \bf{1.57} & \bf{1.43} & 1.36 & \bf{1.60} & \bf{1.41} & 1.29 & 1.57 & \bf{1.37} & 1.36 & \bf{1.51} & \bf{1.42} & 1.28 \\
Fanar-1-9B (Random) & 1.46 & 1.40 & \bf{1.32} & 1.50 & 1.38 & \bf{1.37} & 1.54 & 1.39 & \bf{1.37} & 1.54 & 1.36 & \bf{1.45} & 1.46 & 1.37 & \bf{1.38} \\
\midrule
LLaMA-3-8B  & 1.50 & 1.18 & 1.06 & 1.62 & 1.19 & 1.09 & 1.50 & 1.19 & 1.05 & 1.48 & 1.18 & 1.12 & 1.41 & 1.20 & 1.07 \\
LLaMA-3-8B (Curriculum) & 1.82 & 1.97 & \bf{1.59} & \bf{1.90} & 1.92 & \bf{1.70} & \bf{1.88} & \bf{1.97} & 1.70 & 1.88 & 1.88 & \bf{1.78} & 1.80 & 1.85 & 1.59 \\
LLaMA-3-8B (Random) & \bf{1.90} & \bf{2.06} & \bf{1.59} & 1.89 & \bf{1.95} & 1.67 & 1.87 & \bf{1.97} & \bf{1.72} & \bf{1.98} & \bf{2.00} & 1.74 & \bf{1.82} & \bf{1.96} & \bf{1.64} \\
\midrule
Qwen-3-8B  & \bf{1.84} & 1.41 & 1.38 & \bf{2.24} & 1.38 & 1.34 & \bf{1.91} & 1.35 & 1.49 & 2.02 & 1.39 & 1.35 & 1.75 & 1.35 & 1.38 \\
Qwen-3-8B (Curriculum) & 1.76 & 1.98 & 1.51 & 1.77 & 1.94 & \bf{1.57} & 1.84 & 1.97 & 1.53 & 1.79 & 1.90 & \bf{1.64} & 1.75 & 1.94 & \bf{1.58} \\
Qwen-3-8B (Random) & 1.79 & \bf{2.05} & \bf{1.52} & 1.84 & \bf{2.01} & 1.55 & 1.82 & \bf{1.99} & \bf{1.58} &\bf{ 1.80} & \bf{1.98} & 1.63 & \bf{1.76} & \bf{1.96} & 1.51 \\

\bottomrule
\end{tabular}}
\caption{LLM-as-a-Judge evaluation across models, training types, and Arabic dialects. We use the following notation: G = Generation, C = Continuation, R = Revision.}
\label{tab:results}
\end{table*}

%% file: tables/results_aspect.tex
\begin{table*}[t]
\centering
\small
\setlength{\tabcolsep}{3pt}
\resizebox{\textwidth}{!}{
\begin{tabular}{lccccccccccccccc}
\toprule
\multirow{2}{*}{\textbf{Model}} & \multicolumn{5}{c}{\textbf{Generation}} & \multicolumn{5}{c}{\textbf{Continuation}} & \multicolumn{5}{c}{\textbf{Revision}} \\
\cmidrule(lr){2-6} \cmidrule(lr){7-11} \cmidrule(lr){12-16}
& Comp & Flue & Cohe & Poet & Over & Comp & Flue & Cohe & Poet & Over & Comp & Flue & Cohe & Poet & Over \\
\midrule
ALLaM-7B-Instruct & 1.96 & 2.09 & 2.14 & 1.70 & 1.97 & 1.48 & 1.46 & 1.33 & 1.27 & 1.38 & 1.14 & 1.41 & 1.31 & 1.25 & 1.28 \\
ALLaM-7B-Instruct (Cur) & 2.45 & \bf{2.79} & \bf{2.78} & \bf{2.29} & \bf{2.58} & 2.14 & 2.06 & 1.83 & 1.76 & 1.95 & \bf{1.43} & \bf{1.84} & \bf{1.64} & \bf{1.58} & \bf{1.62} \\
ALLaM-7B-Instruct (Rand) & \bf{2.46} & 2.77 & 2.75 & 2.28 & 2.56 & \bf{2.20} & \bf{2.10} & \bf{1.89} & \bf{1.80} & \bf{2.00} & \bf{1.43} & 1.81 & 1.63 & 1.57 & 1.61 \\
\midrule
Fanar-1-9B & 1.04 & \bf{1.91} & 1.34 & 1.02 & 1.40 & 1.07 & 1.08 & 1.07 & 1.04 & 1.06 & 1.04 & 1.11 & 1.11 & 1.05 & 1.08 \\
Fanar-1-9B (Cur) & 1.73 & 1.61 & \bf{1.36} & \bf{1.51} & 1.55 & \bf{1.63} & \bf{1.44} & \bf{1.26} & \bf{1.30} & \bf{1.41} & 1.25 & 1.42 & 1.28 & 1.27 & 1.30 \\
Fanar-1-9B (Rand) & \bf{1.66} & 1.55 & 1.32 & 1.46 & 1.50 & 1.61 & 1.41 & 1.24 & 1.27 & 1.38 & \bf{1.30} & \bf{1.50} & \bf{1.35} & \bf{1.36} & \bf{1.38} \\
\midrule
Llama-3-8B & 1.48 & 1.45 & 1.80 & 1.30 & 1.51 & 1.23 & 1.19 & 1.26 & 1.06 & 1.19 & 1.05 & 1.10 & 1.10 & 1.05 & 1.08 \\
Llama-3-8B (Cur) & 2.04 & 1.99 & 1.70 & 1.71 & 1.86 & 2.23 & 2.03 & 1.71 & 1.71 & 1.92 & 1.51 & 1.87 & \bf{1.68} & \bf{1.62} & \bf{1.67} \\
Llama-3-8B (Rand) & \bf{2.08} & \bf{2.01} & \bf{1.72} & \bf{1.74} & \bf{1.89} & \bf{2.32} & \bf{2.09} & \bf{1.78} & \bf{1.77} & \bf{1.99} & \bf{1.53} & \bf{1.88} & \bf{1.68} & 1.60 & 1.67 \\
\midrule
Qwen-3-8B & \bf{2.01} & \bf{1.94} & \bf{2.23} & \bf{1.68} & \bf{1.96} & 1.44 & 1.45 & 1.40 & 1.22 & 1.38 & 1.28 & 1.45 & 1.50 & 1.34 & 1.39 \\
Qwen-3-8B (Cur) & 1.96 & 1.89 & 1.62 & 1.65 & 1.78 & 2.21 & 2.06 & 1.76 & 1.75 & 1.94 & \bf{1.43} & \bf{1.76} & \bf{1.54} & \bf{1.51} & \bf{1.56}\\
Qwen-3-8B (Rand) & 1.96 & 1.93 & 1.67 & 1.66 & 1.80 & \bf{2.24} & \bf{2.14} & \bf{1.83} & \bf{1.79} & \bf{2.00} & \bf{1.43} & \bf{1.76} & \bf{1.54} & \bf{1.51} & \bf{1.56} \\
\bottomrule
\end{tabular}}
\caption{Results across models and tasks. We use the following notation: Comp = Compliance, Flue = Fluency, Cohe = Coherence, Poet = Poetic Quality, Over = Overall.}
\label{tab:results2}
\end{table*}

%% file: tables/analysis_average.tex
\begin{table}[t]
\centering
\scriptsize
\setlength{\tabcolsep}{6pt}
\renewcommand{\arraystretch}{1.1}

\begin{tabular}{l c}
\toprule
\textbf{Model} & \textbf{Analysis Accuracy} \\
\midrule
\allam                & 66.2 \\
\allam\,(Curriculum)  & 77.8 \\
\allam\,(Random)      & \textbf{78.1} \\
\midrule
\fanar                & 51.2 \\
\fanar\,(Curriculum)  & \textbf{60.5} \\
\fanar\,(Random)      & 60.1 \\
\midrule
\llama                & 47.0 \\
\llama\,(Curriculum)  & 78.7 \\
\llama\,(Random)      & \textbf{79.0} \\
\midrule
\qwen                 & 44.6 \\
\qwen\,(Curriculum)   & \textbf{77.4} \\
\qwen\,(Random)       & 77.2 \\
\bottomrule
\end{tabular}

\caption{Analysis task accuracy (\%). \textbf{Bold} indicates the best result within each model family. The results are averaged over all subtasks.}
\label{tab:analysis_only}
\end{table}

%% file: tables/human_evaluation.tex
\begin{table}[t]
\centering

\small

\begin{tabularx}{0.98\columnwidth}{lXXXXX}
\toprule
\textbf{Model} & \textbf{Comp} & \textbf{Flue} & \textbf{Cohe} & \textbf{Poet} & \textbf{Over} \\
\midrule

ALLaM-7B (Base) & 3.09 & 3.10 & 3.02 & 2.87 & 3.02 \\
ALLaM-7B (FT) & \textbf{3.93} & \textbf{4.20} & \textbf{4.00} & \textbf{3.82} & \textbf{3.99} \\
\midrule
Qwen3-8B (Base) & 2.48 & 2.13 & 2.46 & 1.91 & 2.24 \\
Qwen3-8B (FT) & \textbf{3.66} & \textbf{3.86} & \textbf{3.54} & \textbf{3.58} & \textbf{3.66} \\
\bottomrule
\end{tabularx}
\caption{Human evaluation results averaged across two annotators. Scores are on a 1--5 scale (higher is better). All pairwise differences between models are statistically significant ($p < 0.0001$). Comp = Compliance, Flue = Fluency, Cohe = Coherence, Poet = Poetic Quality, Over = Overall.}
\label{tab:model_performance}
\end{table}

%% file: tables/ift_data_stats.tex
\begin{table}[tbh]
\centering
\small
\begin{tabular}{lrr}
\toprule
\textbf{Value} & \textbf{Count} & \textbf{\%} \\
\midrule
\multicolumn{3}{l}{\textbf{Meter}} \\
\textarabic{بحر الطويل} (Al-Ṭawīl) & 15164 & 20.31 \\
\textarabic{بحر الكامل} (Al-Kāmil) & 12017 & 16.10 \\
\textarabic{بحر البسيط} (Al-Basīṭ) & 9162 & 12.27 \\
\textarabic{بحر الوافر} (Al-Wāfir) & 6376 & 8.54 \\
\textarabic{بحر الخفيف} (Al-Khafīf) & 5553 & 7.44 \\
\textarabic{بحر السريع} (As-Sarīʿ) & 3544 & 4.75 \\
\textarabic{بحر الرجز} (Ar-Rajaz) & 3298 & 4.42 \\
\textarabic{بحر المتقارب} (Al-Mutaqārib) & 3002 & 4.02 \\
\textarabic{بحر الرمل} (Ar-Raml) & 2905 & 3.89 \\
\textarabic{بحر المجتث} (Al-Mujtath) & 1897 & 2.54 \\
\midrule
\multicolumn{3}{l}{\textbf{Poet Era}} \\
\textarabic{العصر الحديث} (Modern) & 82487 & 37.17 \\
\textarabic{العصر العباسي} (Abbasid) & 50125 & 22.59 \\
\textarabic{العصر المملوكي} (Mamluk) & 22298 & 10.05 \\
\textarabic{العصر العثماني} (Ottoman) & 17011 & 7.66 \\
\textarabic{العصر الأندلسي} (Andalusian) & 13507 & 6.09 \\
\textarabic{العصر الأموي} (Umayyad) & 9207 & 4.15 \\
\textarabic{العصر الفاطمي} (Fatimid) & 6893 & 3.11 \\
\textarabic{العصر الأيوبي} (Ayyubid) & 5736 & 2.58 \\
\textarabic{المخضرمون} (Mukhaḍramūn) & 5477 & 2.47 \\
\textarabic{العصر الجاهلي} (Pre-Islamic) & 4906 & 2.21 \\

\midrule
\multicolumn{3}{l}{\textbf{Genre}} \\
\textarabic{عامه} (General) & 14139 & 22.89 \\
\textarabic{قصيره} (Short) & 7171 & 11.61 \\
\textarabic{مدح} (Praise) & 4947 & 8.01 \\
\textarabic{رومانسيه} (Romantic) & 4442 & 7.19 \\
\textarabic{هجاء} (Satire) & 2923 & 4.73 \\
\textarabic{حزينه} (Sad) & 2163 & 3.50 \\
\textarabic{عتاب} (Reproach) & 1575 & 2.55 \\
\textarabic{دينيه} (Religious) & 1244 & 2.01 \\
\textarabic{رثاء} (Elegy) & 1239 & 2.01 \\
\textarabic{غزل} (Ghazal) & 1189 & 1.93 \\

\bottomrule
\end{tabular}
\caption{Top 10 most frequent values for poetic meter, poet era, and genre.}
\label{tab:corpus_stats}
\end{table}

\begin{table*}[!t]
\centering
\scriptsize
\setlength{\tabcolsep}{4pt}
\begin{tabularx}{\textwidth}{Xr}
\toprule
\textbf{Subtask (Input $\rightarrow$ Output)} & \textbf{Samples Count} \\
\midrule
\multicolumn{2}{c}{\textit{Train Split}} \\
\midrule
poem\_text $\rightarrow$ poet\_name & 142,583 \\
poem\_text $\rightarrow$ poem\_title & 88,464 \\
poem\_text $\rightarrow$ keywords & 58,700 \\
poem\_text $\rightarrow$ poet\_era & 32,892 \\
poet\_name $\rightarrow$ poet\_era & 26,142 \\
poet\_name, poem\_text $\rightarrow$ poet\_era & 18,055 \\
poet\_name, poem\_text $\rightarrow$ rhyme & 15,389 \\
poem\_text $\rightarrow$ meter & 11,168 \\
poem\_text $\rightarrow$ genre & 6,331 \\
poet\_name, $\rightarrow$ meter & 6,306 \\
poet\_name, poem\_text $\rightarrow$ meter & 5,132 \\
poet\_name $\rightarrow$ genre & 4,843 \\
poet\_name, poem\_text $\rightarrow$ genre & 3,889 \\
poet\_name, poem\_text, genre $\rightarrow$ meter & 1,241 \\
\midrule
\multicolumn{2}{c}{\textit{Test Split}} \\
\midrule
poem\_text $\rightarrow$ poet\_name & 1,042 \\
poem\_text $\rightarrow$ poem\_title & 876 \\
poem\_text $\rightarrow$ meter & 745 \\
poem\_text $\rightarrow$ poet\_era & 638 \\
poem\_text $\rightarrow$ keywords & 621 \\
poem\_text $\rightarrow$ genre & 550 \\
poet\_name $\rightarrow$ poet\_era & 476 \\
poet\_name $\rightarrow$ genre & 416 \\
poet\_name $\rightarrow$ meter & 364 \\
poet\_name, poem\_text $\rightarrow$ genre & 321 \\
poet\_name, poem\_text $\rightarrow$ meter & 284 \\
poet\_name, poem\_text $\rightarrow$ rhyme & 226 \\
poet\_name, poem\_text $\rightarrow$ poet\_era & 225 \\
poet\_name, poem\_text, genre $\rightarrow$ meter & 200 \\
\bottomrule
\end{tabularx}
\caption{Detailed statistics per subtask for the \textit{Analysis} task in the Arabic poetry IFT dataset.}
\label{tab:ift_analysis_subtasks}
\end{table*}

\begin{table*}[!t]
\centering
\scriptsize
\setlength{\tabcolsep}{4pt}
\begin{tabularx}{\textwidth}{Xr}
\toprule
\textbf{Subtask (Input $\rightarrow$ Output)} & \textbf{Samples Count} \\
\midrule
\multicolumn{2}{c}{\textit{Train Split}} \\
\midrule
poem\_title, $\rightarrow$ poem\_text & 99,852 \\
poet\_name, $\rightarrow$ poem\_text & 85,765 \\
poem\_title, poet\_name $\rightarrow$ poem\_text & 52,501 \\
keywords, $\rightarrow$ poem\_text & 45,128 \\
key\_phrases, $\rightarrow$ poem\_text & 34,584 \\
poet\_name, poet\_era $\rightarrow$ poem\_text & 21,509 \\
rhyme, $\rightarrow$ poem\_text & 18,398 \\
poet\_era, poem\_title $\rightarrow$ poem\_text & 16,781 \\
meter, $\rightarrow$ poem\_text & 11,761 \\
poet\_name, rhyme $\rightarrow$ poem\_text & 10,290 \\
poem\_title, rhyme $\rightarrow$ poem\_text & 7,594 \\
poet\_name, meter $\rightarrow$ poem\_text & 6,879 \\
genre, $\rightarrow$ poem\_text & 6,067 \\
poet\_name, genre $\rightarrow$ poem\_text & 3,142 \\
poem\_title, genre $\rightarrow$ poem\_text & 2,378 \\
rhyme, meter $\rightarrow$ poem\_text & 1,348 \\
genre, poet\_era $\rightarrow$ poem\_text & 1,337 \\
poem\_title, meter $\rightarrow$ poem\_text & 1,226 \\
genre, meter $\rightarrow$ poem\_text & 797 \\
\midrule
\multicolumn{2}{c}{\textit{Test Split}} \\
\midrule
poem\_title $\rightarrow$ poem\_text & 849 \\
poet\_name $\rightarrow$ poem\_text & 728 \\
meter $\rightarrow$ poem\_text & 630 \\
genre $\rightarrow$ poem\_text & 549 \\
keywords $\rightarrow$ poem\_text & 513 \\
key\_phrases $\rightarrow$ poem\_text & 443 \\
rhyme $\rightarrow$ poem\_text & 430 \\
poem\_title, poet\_name $\rightarrow$ poem\_text & 428 \\
poet\_era, poem\_title $\rightarrow$ poem\_text & 376 \\
poet\_name, poet\_era $\rightarrow$ poem\_text & 333 \\
poet\_name, meter $\rightarrow$ poem\_text & 295 \\
poet\_name, genre $\rightarrow$ poem\_text & 262 \\
poem\_title, meter $\rightarrow$ poem\_text & 210 \\
poet\_name, rhyme $\rightarrow$ poem\_text & 208 \\
poem\_title, genre $\rightarrow$ poem\_text & 187 \\
poem\_title, rhyme $\rightarrow$ poem\_text & 148 \\
genre, poet\_era $\rightarrow$ poem\_text & 139 \\
rhyme, meter $\rightarrow$ poem\_text & 131 \\
genre, meter $\rightarrow$ poem\_text & 125 \\
\bottomrule
\end{tabularx}
\caption{Detailed statistics per subtask for the \textit{Generation} task in the Arabic poetry IFT dataset.}
\label{tab:ift_generation_subtasks}
\end{table*}

\begin{table*}[!t]
\centering
\scriptsize
\setlength{\tabcolsep}{4pt}
\begin{tabularx}{\textwidth}{Xr}
\toprule
\textbf{Subtask (Input $\rightarrow$ Output)} & \textbf{Samples Count} \\
\midrule
\multicolumn{2}{c}{\textit{Train Split}} \\
\midrule
existing\_verses, poem\_title $\rightarrow$ poem\_continuation & 138,055 \\
existing\_verses, poem\_title, poet\_name $\rightarrow$ poem\_continuation & 77,174 \\
existing\_verses, keywords $\rightarrow$ poem\_continuation & 64,862 \\
existing\_verses, poet\_era $\rightarrow$ poem\_continuation & 49,948 \\
existing\_verses, meter $\rightarrow$ poem\_continuation & 25,602 \\
existing\_verses, rhyme $\rightarrow$ poem\_continuation & 24,826 \\
existing\_verses, poem\_title, poet\_era $\rightarrow$ poem\_continuation & 19,147 \\
existing\_verses, poem\_title, rhyme $\rightarrow$ poem\_continuation & 12,453 \\
existing\_verses, genre $\rightarrow$ poem\_continuation & 8,082 \\
existing\_verses, poem\_title, genre $\rightarrow$ poem\_continuation & 4,164 \\
existing\_verses, poem\_title, meter $\rightarrow$ poem\_continuation & 2,963 \\
\midrule
\multicolumn{2}{c}{\textit{Test Split}} \\
\midrule
existing\_verses, poem\_title $\rightarrow$ poem\_continuation & 1,218 \\
existing\_verses, meter $\rightarrow$ poem\_continuation & 999 \\
existing\_verses, poet\_era $\rightarrow$ poem\_continuation & 833 \\
existing\_verses, keywords $\rightarrow$ poem\_continuation & 730 \\
existing\_verses, genre $\rightarrow$ poem\_continuation & 702 \\
existing\_verses, rhyme $\rightarrow$ poem\_continuation & 536 \\
existing\_verses, poem\_title, meter $\rightarrow$ poem\_continuation & 521 \\
existing\_verses, poem\_title, poet\_name $\rightarrow$ poem\_continuation & 449 \\
existing\_verses, poem\_title, poet\_era $\rightarrow$ poem\_continuation & 391 \\
existing\_verses, poem\_title, genre $\rightarrow$ poem\_continuation & 340 \\
existing\_verses, poem\_title, rhyme $\rightarrow$ poem\_continuation & 265 \\
\bottomrule
\end{tabularx}
\caption{Detailed statistics per subtask for the \textit{Continuation} task in the Arabic poetry IFT dataset.}
\label{tab:ift_continuation_subtasks}
\end{table*}

\begin{table*}[!t]
\centering
\scriptsize
\setlength{\tabcolsep}{4pt}
\begin{tabularx}{\textwidth}{Xr}
\toprule
\textbf{Subtask (Corruption Type)} & \textbf{Samples Count} \\
\midrule
\multicolumn{2}{c}{\textit{Train Split}} \\
\midrule
rhyme\_structure & 11,196 \\
full\_style & 9,896 \\
rhyme\_substitution & 9,834 \\
rhyme\_content & 9,822 \\
era\_corruption & 8,915 \\
meter\_transformation & 7,294 \\
meter\_destruction & 7,292 \\
meter\_inconsistency & 4,698 \\
\midrule
\multicolumn{2}{c}{\textit{Test Split}} \\
\midrule
meter\_destruction & 483 \\
rhyme\_structure & 483 \\
rhyme\_content & 483 \\
meter\_inconsistency & 483 \\
rhyme\_substitution & 483 \\
era\_corruption & 483 \\
meter\_transformation & 483 \\
full\_style & 482 \\
\bottomrule
\end{tabularx}
\caption{Detailed statistics per subtask for the \textit{Corruption (Restoration)} task in the Arabic poetry IFT dataset.}
\label{tab:ift_corruption_subtasks}
\end{table*}

%% file: tables/IFT_tempelates.tex
\begin{table*}[t]
\centering
\scriptsize
\begin{tabularx}{\textwidth}{@{}l l l *5{X}@{}}
\toprule
\textbf{Task} & \textbf{Input} & \textbf{Output} &
\textbf{MSA} & \textbf{Nile Valley} & \textbf{North Africa} &
\textbf{Gulf} & \textbf{Levant} \\
\midrule

\textbf{Generation}
& \shortstack{Poet Name\\Era}
& Poem Text
& \textarabic{اكتب قصيدة تحاكي أسلوب ((poet\_name)) من زمن ((poet\_era)).}
& \textarabic{اكتب قصيدة تقلد أسلوب ((poet\_name)) من زمن ((poet\_era)).}
& \textarabic{كتب قصيدة بالأسلوب ديال ((poet\_name)) من زمان ((poet\_era)).}
& \textarabic{اكتب قصيدة على أسلوب ((poet\_name)) من زمن ((poet\_era)).}
& \textarabic{اكتب قصيدة بنفس أسلوب ((poet\_name)) من زمن ((poet\_era)).} \\
\midrule

\textbf{Continuation}
& \shortstack{Existing Verses\\Meter}
& Poem Text
& \textarabic{تابع هذه الأبيات بنفس الوزن الشعري ((meter)). ((existing\_verses))}
& \textarabic{كمل الأبيات دي بنفس الوزن الشعري ((meter)): ((existing\_verses))}
& \textarabic{كمّل هاد لْبيوت بنفس الوزن الشعري ((meter)). ((existing\_verses))}
& \textarabic{كمل هالأبيات هذي بنفس الوزن الشعري ((meter)). ((existing\_verses))}
& \textarabic{كمّل هالأبيات بنفس الوزن الشعري ((meter)). ((existing\_verses))} \\
\midrule

\textbf{Analysis}
& \shortstack{Poet Name\\Poem Text\\Genre}
& Meter
& \textarabic{هذه القصيدة كتبها ((poet\_name)) وتنتمي إلى نوع ((genre)): ((poem\_text)) ما هو البحر الشعري المستخدم؟}
& \textarabic{القصيدة دي كتبها ((poet\_name)) ودي من نوع ((genre)): ((poem\_text)) إيه هو البحر الشعري المستخدم؟}
& \textarabic{هاد القصيدة كتبها ((poet\_name)) وكتنتمي لنوع ((genre)): ((poem\_text)) شنو هو البحر الشعري اللي مستعمل؟}
& \textarabic{هذي القصيدة كتبها ((poet\_name)) ونوعها ((genre)): ((poem\_text)) وشو البحر الشعري المستخدم؟}
& \textarabic{هي القصيدة كتبها ((poet\_name)) وبتنتمي لنوع ((genre)): ((poem\_text)) شو هو البحر الشعري المستخدم؟} \\
\midrule

\textbf{Restoration}
& \shortstack{Poem Text\\Era\\Poet Name\\Meter}
& Poem Text
& \textarabic{لقد تم تدمير البحر الشعري لهذه القصيدة. مهمتك هي إعادة النص الشعري إلى وزنه الأصلي المنتظم ((meter)). أعد كتابة الأبيات لتطابق البحر ((meter)) مع الحفاظ على المعنى والقافية. الشاعر: ((poet\_name)) العصر: ((poet\_era)) البحر: ((meter)) القصيدة المحرفة: ((poem\_text)) القصيدة المستعادة:}
& \textarabic{البحر الشعري للقصيدة دي اتدمر. مهمتك إنك ترجع النص الشعري لوزنه الأصلي ((meter)). اكتب الأبيات بما يطابق البحر ((meter)) مع الحفاظ على المعنى والقافية. الشاعر: ((poet\_name)) العصر: ((poet\_era)) البحر: ((meter)) القصيدة المحرفة: ((poem\_text)) القصيدة المستعادة:}
& \textarabic{تخرب البحر ديال هاد القصيدة. خاصك ترجع النص الشعري للوزن الأصلي ((meter)). كتب الأبيات باش يطابقو البحر ((meter)) وحافظ على المعنى والقافية. الشاعر: ((poet\_name)) العصر: ((poet\_era)) البحر: ((meter)) القصيدة المبدلة: ((poem\_text)) القصيدة المصلحة:}
& \textarabic{البحر الشعري لهالقصيدة تخرب. مهمتك إنك ترجع النص الشعري لوزنه الأصلي ((meter)). عيد كتابة الأبيات لتطابق البحر ((meter)) وحافظ على المعنى والقافية. الشاعر: ((poet\_name)) العصر: ((poet\_era)) البحر: ((meter)) القصيدة المحرفة: ((poem\_text)) القصيدة المستعادة:}
& \textarabic{البحر الشعري لهي القصيدة تخرب. مهمتك ترجع النص لوزنه الأصلي ((meter)). كتوب الأبيات لتتطابق مع البحر ((meter)) مع الحفاظ على المعنى والقافية. الشاعر: ((poet\_name)) العصر: ((poet\_era)) البحر: ((meter)) القصيدة المحرفة: ((poem\_text)) القصيدة المستعادة:} \\
\bottomrule
\end{tabularx}
\caption{Instruction template examples for IFT tasks across MSA and regional Arabic dialects.}
\label{tab:poetry_ift_templates}
\end{table*}

%% file: tables/detailed_analysis_task.tex
\begin{table*}[!t]
\centering
\scriptsize
\setlength{\tabcolsep}{3pt}
\renewcommand{\arraystretch}{1.12}

\resizebox{\textwidth}{!}{%
\begin{tabular}{l r *{13}{r}}
\toprule
\multirow{2}{*}{\textbf{Model}} &
\multirow{2}{*}{\textbf{Joint}} &
\multicolumn{6}{c}{\textbf{Poem-text}} &
\multicolumn{4}{c}{\textbf{Poem+Poet}} &
\multicolumn{3}{c}{\textbf{Poet-name}} \\
\cmidrule(lr){3-8}\cmidrule(lr){9-12}\cmidrule(lr){13-15}
& &
\textbf{G} & \textbf{K} & \textbf{M} & \textbf{Title} & \textbf{Era} & \textbf{Poet} &
\textbf{G} & \textbf{M} & \textbf{Era} & \textbf{Rhy} &
\textbf{G} & \textbf{M} & \textbf{Era} \\
\midrule
\allam & 79.5 & 35.6 & 72.0 & 85.8 & 99.9 & 48.0 & 72.3 & 36.4 & 77.8 & 63.6 & 25.2 & 26.7 & 49.7 & 88.0 \\
\allam(Curriculum) & 99.5 & 29.8 & 94.7 & 99.7 & 99.9 & 66.1 & 87.9 & 29.3 & 99.7 & 79.6 & 98.2 & 30.0 & 54.7 & 88.9 \\
\allam(Random) & 99.5 & 32.9 & 95.3 & 99.6 & 99.9 & 61.9 & 87.8 & 33.6 & 99.7 & 76.9 & 99.6 & 31.0 & 58.2 & 89.3 \\
\midrule
\fanar & 40.5 & 28.9 & 65.5 & 38.4 & 99.9 & 38.4 & 60.3 & 24.6 & 41.5 & 46.7 & 21.2 & 26.0 & 49.2 & 53.8 \\
\fanar(Curriculum) & 59.5 & 33.8 & 92.6 & 48.0 & 99.8 & 48.3 & 60.0 & 33.3 & 66.9 & 59.1 & 22.6 & 28.1 & 57.7 & 78.4 \\
\fanar(Random) & 59.5 & 32.4 & 93.9 & 52.2 & 99.9 & 37.2 & 60.3 & 30.5 & 60.9 & 60.0 & 29.6 & 32.7 & 55.5 & 78.6 \\
\midrule
\llama & 38.5 & 34.7 & 61.7 & 35.4 & 99.9 & 28.5 & 52.3 & 29.0 & 35.9 & 40.4 & 28.8 & 22.4 & 35.2 & 40.6 \\
\llama(Curriculum) & 99.0 & 39.3 & 92.4 & 98.3 & 100.0 & 64.3 & 84.6 & 40.2 & 98.2 & 88.0 & 99.6 & 37.0 & 53.9 & 89.9 \\
\llama(Random) & 98.0 & 38.5 & 93.7 & 98.4 & 100.0 & 63.3 & 86.3 & 40.5 & 98.2 & 85.8 & 96.5 & 36.5 & 56.3 & 91.6 \\
\midrule
\qwen & 45.5 & 22.7 & 49.3 & 43.6 & 99.8 & 17.1 & 54.1 & 28.4 & 46.5 & 29.8 & 26.6 & 23.6 & 31.0 & 33.8 \\
\qwen(Curriculum) & 98.0 & 31.6 & 93.6 & 98.9 & 99.9 & 62.2 & 84.9 & 30.8 & 100.0 & 83.6 & 100.0 & 35.8 & 54.1 & 87.4 \\
\qwen(Random) & 99.5 & 29.6 & 92.6 & 99.1 & 99.9 & 62.4 & 85.4 & 30.5 & 100.0 & 85.3 & 100.0 & 31.7 & 58.5 & 85.7 \\
\bottomrule
\end{tabular}%
}

\caption{Analysis task evaluation results in \% (higher is better). \textbf{Joint} corresponds to \texttt{poet\_name, poem\_text, genre $\rightarrow$ meter}. Abbreviations: G=genre, K=keywords, M=meter, Rhy=rhyme. Models with (Curriculum)/(Random) are LoRA fine-tuned. Rows are grouped by backbone; horizontal rules separate groups.}
\label{tab:multitask-poetry}
\end{table*}

%% file: tables/automatic_metrics.tex
\begin{table*}[t]
\centering
\scriptsize
\setlength{\tabcolsep}{3pt}
\renewcommand{\arraystretch}{1.12}

\begin{tabularx}{\textwidth}{l*{10}{>{\centering\arraybackslash}X}}
\toprule
\multirow{2}{*}{\textbf{Model}} &
\multicolumn{3}{c}{\textbf{Resoration}} &
\multicolumn{4}{c}{\textbf{Generation}} &
\multicolumn{3}{c}{\textbf{Continuation}} \\
\cmidrule(lr){2-4}\cmidrule(lr){5-8}\cmidrule(lr){9-11}
& \textbf{BERT-Score} & \textbf{Rouge-L} & \textbf{Rhyme}
& \textbf{BERT-Score} & \textbf{Rouge-L} & \textbf{Rhyme} & \textbf{Key-Phrase}
& \textbf{BERT-Score} & \textbf{Rouge-L} & \textbf{Rhyme} \\
\midrule

\allam
& 83.3 & 21.9 & 21.5
& 80.6 & 5.9 & 14.4 & \textbf{37.3}
& 84.8 & 4.4 & 52.0 \\
\allam(Curriculum)
& \textbf{86.9} & 30.1 & 55.2
& 80.8 & \textbf{6.9} & 54.5 & 31.7
& \textbf{86.9} & \textbf{4.5} & 78.8 \\
\allam(random)
& \textbf{86.9} & \textbf{30.3} & \textbf{55.8}
& \textbf{80.9} & \textbf{6.9} & \textbf{55.4} & 33.1
& \textbf{86.9} & \textbf{4.5} & \textbf{80.1} \\
\midrule

\fanar
& 76.2 & 7.2 & 14.6
& 74.8 & 3.1 & 7.1 & 21.6
& 77.4 & 4.4 & 24.6 \\
\fanar(Curriculum)
& 81.0 & 20.5 & 44.1
& 78.7 & \textbf{6.2} & 44.3 & 22.1
& \textbf{82.3} & \textbf{10.6} & \textbf{75.0} \\
\fanar(Random)
& \textbf{81.3} & \textbf{22.8} & \textbf{44.7}
& \textbf{78.9} & 6.1 & \textbf{45.3} & \textbf{22.4}
& \textbf{82.3} & 10.5 & 70.4 \\
\midrule

\llama
& 83.5 & \textbf{40.5} & 52.5
& 74.4 & 5.6 & 22.4 & \textbf{61.0}
& 80.8 & \textbf{4.6} & 50.5 \\
\llama(Curriculum)
& 85.9 & 32.5 & 64.0
& \textbf{81.6} & 6.2 & 52.5 & 38.53
& \textbf{86.5} & 4.5 & 85.2 \\
\llama(Random)
& \textbf{86.1} & 33.6 & \textbf{66.7}
& 81.5 & \textbf{6.3} & \textbf{53.5} & 37.3
& \textbf{86.5} & 4.4 & \textbf{86.3} \\
\midrule

\qwen
& \textbf{86.9} & \textbf{47.0} & 31.1
& \textbf{80.8} & \textbf{6.4} & 11.3 & \textbf{63.8}
& 84.3 & \textbf{4.4} & 35.4 \\
\qwen(Curriculum)
& 83.8 & 27.8 & \textbf{64.2}
& 79.7 & 5.9 & 47.1 & \textbf{63.8}
& \textbf{85.2} & 4.3 & 82.2 \\
\qwen(Random)
& 83.9 & 28.6 & \textbf{64.2}
& 79.6 & 6.0 & \textbf{50.9} & \textbf{63.8}
& 84.9 & 4.3 & \textbf{83.5} \\

\bottomrule
\end{tabularx}

\caption{Automatic evaluation results in \% (higher is better). \textbf{Bold} indicates the best score within each model family (ties are bolded).}
\label{tab:auto_eval}

\end{table*}